\documentclass[letterpaper, 10 pt, conference]{ieeeconf}
\usepackage[T1]{fontenc}
\IEEEoverridecommandlockouts
\usepackage{amsmath,amssymb,graphicx,epstopdf,enumerate,booktabs,setspace,float,stfloats,bm, multirow, lscape, color,soul, subcaption, url, hyperref}

\usepackage[bottom=2cm,top=2.01cm,left=1.69cm,right=1.69cm]{geometry}

\DeclareMathOperator{\diffv}{\emph{diff}}
\usepackage[normalem]{ulem}
\usepackage{soul}   
\usepackage{color}  
\usepackage{xcolor} 
\usepackage[noadjust]{cite}
\usepackage{aecompl}

\soulregister\ref7
\soulregister\cite7
\soulregister\eqref7
\usepackage{multirow}
\usepackage{makecell}
\usepackage{booktabs} 

\usepackage[linesnumbered,ruled,vlined]{algorithm2e}

\title{\LARGE \bf Reinforcement Learned Distributed Multi-Robot Navigation with Reciprocal Velocity Obstacle Shaped Rewards}

\author{Ruihua Han$^{1,2}$, Shengduo Chen$^{2}$, Shuaijun Wang$^{2}$, Zeqing Zhang$^{1}$, Rui Gao$^{2}$, Qi Hao$^{2,\dagger}$, Jia Pan$^{1,\dagger}$
\thanks{This work is partially supported by the Shenzhen Fundamental Research Program (No: JCYJ20200109141622964) and is partially supported by HKSAR RGC GRF 11202119, 11207818, and the Innovation and HKSAR Technology Commission under the InnoHK initiative.}
\thanks{$^\dagger$ denotes the corresponding authors: jpan@cs.hku.hk, haoq@sustc.edu.cn}
\thanks{$^{1}$ the Department of Computer Science, The University of Hong Kong, hanrh@connect.hku.hk}
\thanks{$^{2}$ the Department of Computer Science and Engineering, Southern University of Science and Technology}
}

\begin{document}
\maketitle

\begin{abstract}
The challenges to solving the collision avoidance problem lie in adaptively choosing optimal robot velocities in complex scenarios full of interactive obstacles. In this paper, we propose a distributed approach for multi-robot navigation which combines the concept of reciprocal velocity obstacle (RVO) and the scheme of deep reinforcement learning (DRL) to solve the reciprocal collision avoidance problem under limited information. The novelty of this work is threefold: (1) using a set of sequential VO and RVO vectors to represent the interactive environmental states of static and dynamic obstacles, respectively; (2) developing a bidirectional recurrent module based neural network, which maps the states of a varying number of surrounding obstacles to the actions directly; (3) developing a RVO area and expected collision time based reward function to encourage reciprocal collision avoidance behaviors and trade off between collision risk and travel time. The proposed policy is trained through simulated scenarios and updated by the actor-critic based DRL algorithm. We validate the policy in complex environments with various numbers of differential drive robots and obstacles. The experiment results demonstrate that our approach outperforms the state-of-art methods and other learning based approaches in terms of the success rate, travel time, and average speed. Source code of this approach is available at \href{https://github.com/hanruihua/rl_rvo_nav}{\url{https://github.com/hanruihua/rl_rvo_nav}}.

\end{abstract}

\section{Introduction} \label{section 1}

Multi-robot navigation systems have been widely used in many applications to improve productivity and reduce labor costs. In general, there are two types of multi-robot navigation systems: \emph{centralized} and \emph{distributed} methods. In a centralized system, the controller can flexibly coordinate multiple robots in the same workspace to avoid collisions given complete information about the whole swarm. The commercial centralized planners normally plan paths for robots without considering collisions at first and then employ scheduling schemes to avoid collisions at potential conflict points, such as the crossroad of the planned paths~\cite{zhang2021efficient}. However, it is known that the centralized system requires more computation budget when the number of robots increases, and it may also suffer from the signal delay or instability between each individual robot and the central controller. 

The distributed multi-robot system allows each robot to make decisions independently based on onboard sensors. As such, it is suitable for deploying a large number of robots with a relative low computational budget. One of its major challenges is how to achieve reliable collision avoidance with limited sensing information and determine optimal velocities independently with sufficient safety and high efficiency.

Some approaches design the dynamically-feasible trajectories for each robot within a time interval in real time to achieve collision-free navigation, followed by replanning the new trajectories in the next time horizon~\cite{MPC3}. However, the finer discretization of time requires more computation resources to get more accurate results. Other approaches calculate the optimal velocities directly with low costs, such as the potential field based approaches, where the concept of potential field, including artificial attraction and repulsion, is utilized to find the collision-free and time-efficient velocities~\cite{guerra2016avoiding}. Nevertheless, the problem of handling the local minima is challenging.

Velocity obstacle (VO) based approaches as well as their extensions~\cite{van2011reciprocal, hennes2012multi, alonso2013optimal, han2020distributed} are widely used for dynamic collision avoidance, which predicts collision regions for robots and determines the velocity of each robot to avoid reaching the union of these regions in real time. In contrast, deep reinforcement learning (DRL) based approaches represent environment state with suitable data structure and train the neural networks with manually designed rewards~\cite{lowe2017multi, long2018towards, liang2020crowd, everett2018motion}. DRL approaches are advantageous in converting the rich training experiences into capabilities of taking multiple steps ahead into account and achieving more aggressive movement decisions. 

However, there are three challenges when developing a fully functional DRL based multi-robot navigation system:
\begin{enumerate}[1)]
    \item \textbf{Interactive environment state representation.} How to develop a proper form of environment state representation which can explicitly describe the collision avoidance interactions between those robots is still an open problem.
    \item \textbf{Efficient mapping from sequential environment states to continuous control actions.} How to achieve the optimal and precise control actions for collision avoidance given the current sequential environment state using neural networks at a low computational cost is still challenging.
    \item \textbf{Reward design for collision avoidance behavior regulation.} There is still no systematic method to design rewards according to the observations for representing the collision risk precisely and guiding robots to achieve reciprocal collision avoidance (RCA) behaviors.
\end{enumerate}

Some methods use a set of vectors containing positions and velocities from multiple robots as the policy input~\cite{chen2017socially, han2020cooperative}, which does not directly describe the collision avoidance interaction constraints among robots and hence demands extra capabilities of DRL networks to derive those constraints. To map the state of varying number of robots to actions, some approaches use recurrent neural networks (RNNs) to extract invariant features from the sequential input and output the value function to select optimal actions from a discrete action space~\cite{everett2018motion, everett2021collision}, but the unidirectional RNNs tend to focus on the recent input robot information instead of the information of all those robots. Besides, the high computational cost of the control action limits the applications of these approaches. For the reward design, a dynamic window approach (DWA) is used to encourage those local robots' behaviors towards dynamically feasible areas~\cite{dwa-rl}. However, such a method does not encourage reciprocal collision avoidance behavior among multiple robots.

To this end, we propose a distributed multi-robot navigation approach combining the benefits of the VO concept with the DRL framework to achieve the reciprocal collision avoidance task under limited surrounding information in the shared workspace. Specifically, we utilize VO vectors to model both dynamic agents and static obstacles and employ bidirectional gated recurrent units (GRUs) based neural network for feature extraction and action calculation under the continuous space. The reward function is also designed based on the RVO area. The main contributions of this work include:

\begin{enumerate}[1)]
    \item We represent the information of surrounding robots and obstacles in terms of RVO and VO vectors respectively to achieve a unified environment state representation, which enables each robot to explicitly describe its (reciprocal) collision avoidance interactions with both dynamic agents and static obstacles.
    \item We develop a bidirectional gated recurrent units (BiGRUs) based neural network to extract environment features from the sequential input and map the features to the continuous control action directly with low cost.
    \item We develop a novel reward function based on the RVO or VO areas and expected collision time, which encourages robots to learn reciprocal local collision avoidance behaviors under diverse situations.
\end{enumerate}

Section~\ref{section 2} reviews the related work, followed by the introduction of the system framework and the problem statement in Section~\ref{section 3}. Then, Section~\ref{section 4} describes the deep reinforcement learning network structure, and Section~\ref{section 5} provides the training algorithm in detail. The proposed method is validated by experiments in Section~\ref{section 6}. Finally, the conclusion and future work are presented in Section~\ref{section 7}.

\section{Related work} \label{section 2}

\subsection{Reciprocal Velocity Obstacle}

The central concept of VO is to formulate a potential collision area for a moving obstacle using the relative velocity and position. A velocity outside this area is then chosen for the robot to complete the collision avoidance task. Based on this concept, RVO has been developed to achieve reciprocal collision avoidance (RCA)~\cite{van2008reciprocal}, where each robot has a similar navigation policy and uses an equal effort to avoid each other. RVO has been further developed into optimal reciprocal collision avoidance (ORCA)~\cite{van2011reciprocal}, where several half-plane constraints in the velocity space are used to help the robot find the optimal velocity through linear programming. Non-Holonomic ORCA (NH-ORCA) combines the kinematic model with ORCA to guarantee the ORCA performance under non-holonomic constraints~\cite{alonso2013optimal}. However, the perfect sensing assumption limits the performance of these approaches in real-world applications. Therefore, some approaches combine VO based methods with robot kinematic constraints to account for the uncertainties in localization and control by increasing the collision radius of the robot~\cite{hennes2012multi, claes2018multi}. Other approaches propose a probabilistic variant of RVO (PRVO) to overcome Gaussian uncertainties~\cite{gopalakrishnan2017prvov}. Despite all, most of these VO based approaches are overly conservative, which inevitably reduces navigation efficiency.  

\subsection{Deep Reinforcement Learning}

On the other hand, DRL based collision avoidance approaches can take a large number of training experiences into account and are advantageous in tackling complex scenarios with high efficiency and robustness. Specifically, collision avoidance with DRL (CADRL) employs a value network to generate a collision-free path toward the goal~\cite{chen2017decentralized}. Some approaches focus on the sensor-level collision avoidance policy that maps raw sensor data to the robot control vector with end-to-end training~\cite{long2018towards, Wenzel2021VisionBasedMR, fan2020distributed}.

Compared with sensor-level methods, agent-level DRL approaches use environment models other than raw sensor data to achieve high computational efficiency as well as flexibility for sensor modalities and kinematic/dynamic details~\cite{chen2017socially}. However, the dimension of the input data for a neural network is required to be fixed. Thus, for the environment model with the time-varying number of surrounding robots, some approaches assume that the number of obstacles is a constant and has an upper limit~\cite{han2020cooperative}. RNNs are able to tackle a variable number of moving obstacles, such as in GA3C-CADRL~\cite{everett2018motion}, where the exteroceptive measurements at each step are rearranged as the sequential input data through the long short-term memory (LSTM) module to produce a fixed-size feature vector of the environment. However, such a fixed-size vector likely gives more weights to the final section of the input sequences, which limits the performance in the obstacle-dense environment. Thus, the socially aware RL (SARL) has been proposed~\cite{chen2019crowd}, which utilizes the self-attention mechanism to infer the relative importance of the surrounding dynamic obstacles but without emphasizing the reciprocal collision avoidance interactions and the ability of tackling noncircular obstacles. In this paper, we use RVO vectors as input instead of robot positions and velocities, which can better describe the reciprocal collision interactions among robots and model the obstacles with various shapes. A bidirectional GRUs based neural network is designed to map the state of a varying number of surrounding robots to the continuous action directly with low-cost computation.

Moreover, a proper reward function is the central part of neural policy training. For most DRL based navigation approaches, the reward functions are designed based on the distance among robots, goals, and obstacles, which represent the collision possibility~\cite{Wang2019ResearchOI, long2018towards}. Both behaviors of moving to the goal and keeping away from obstacles can be assigned positive and negative rewards easily. In this paper, the reward function is designed based on the RVO area and using the expected collision time to represent the collision risk, which encourages the RCA regular behaviors.

\section{System Setup and Problem Statement}\label{section 3}

\subsection{System Framework}

In our system, a group of differential drive robots navigates in a shared workspace with a series of positions and velocities along $x$ and $y$ directions. Each robot has no communication with the others but can sense the surrounding robots within a certain range. The information for each robot includes the robot radius, current velocity, current orientation, and desired velocity. The desired velocity is the maximum velocity for the robot moving from the current position to its goal directly without considering obstacles. Similarly, the information of the surrounding robots consists of the robot radiuses, relative positions, and velocities, which can be represented in terms of RVO vectors~\cite{van2008reciprocal}. Moreover, the static line obstacles can also be represented by VO vectors~\cite{snape2011hybrid}. These vectors are fed into a neural network to train the collision avoidance policy. The reward function, which helps evaluate the states and actions, is designed based on the VO area and expected collision time under the current velocities. The navigation policy is trained in simulated scenarios and optimized by using the PPO algorithm. Whereas the policy output of most RL based approaches is the control vector for the robot directly, that of our approach is an increment of the current velocity at each time step, that is, the change rate of the current velocity, which can achieve smoother and efficient robot control. Finally, the output velocity is decomposed into the linear and angular components to control the differential drive robot.

\subsection{Reciprocal Velocity Obstacle}

\begin{figure*}[tp]
    \centering
    \includegraphics[width=0.85\textwidth, clip]{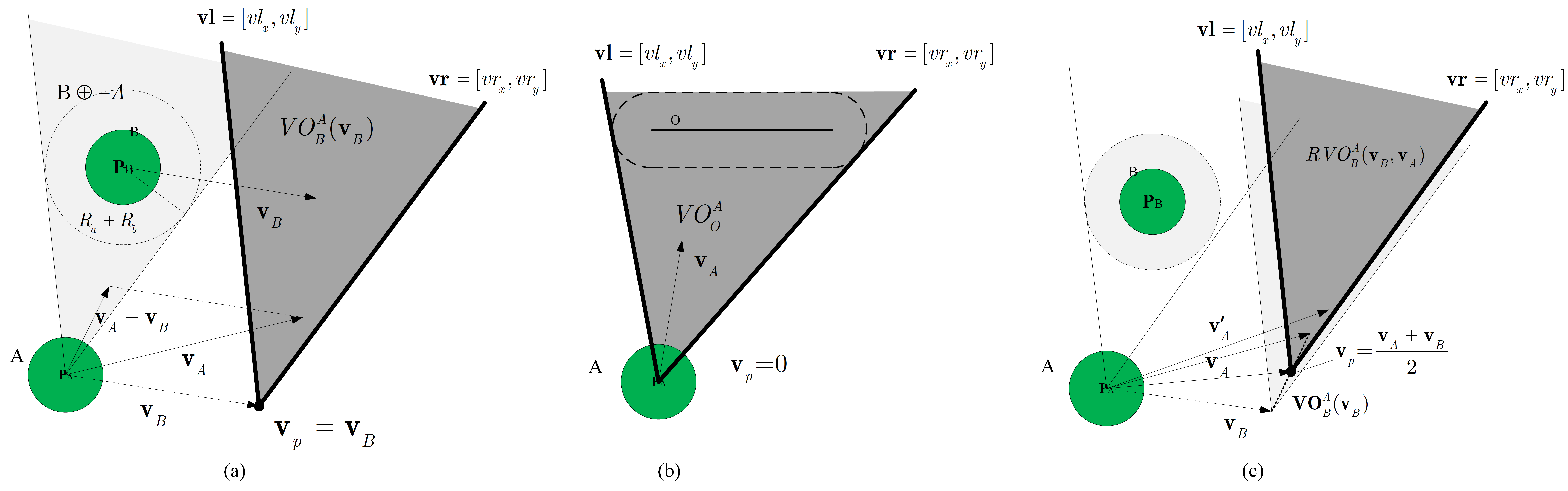}
    \caption{An illustration of velocity obstacle and reciprocal velocity obstacle for the dynamic agent and static line obstacle.}
    \label{rvo}
\end{figure*}

This section summarizes the geometrical definition of the VO and RVO. For disc-shaped robot $A$ and robot $B$ with radiuses ${R_a}$ and ${R_b}$, the positions and velocities can be denoted as ${{\bf{p}}_A}$, ${{\bf{p}}_B}$, ${{\bf{v}}_A}$, ${{\bf{v}}_B}$, respectively. The VO area of the robot $A$ generated by the robot $B$ can be given by the following formula:
\begin{equation}
VO_B^A({{\bf{v}}_B}) = \{ {{\bf{v}}_A}\left| {\lambda ({{\bf{p}}_A},{{\bf{v}}_A} - {{\bf{v}}_B}) \cap B \oplus  - A \ne \emptyset } \right.\}, 
\end{equation}
where $\lambda ({\bf{p}},{\bf{v}})$ denotes the ray with a starting point of ${\bf{p}}$ and in the direction of ${\bf{v}}$, and $ \oplus $ is the Minkowski sum. As shown in Fig.~\ref{rvo} (a), a VO area can be constructed by an apex, where ${{\bf{v}}_p}{\rm{ = }}{{\bf{v}}_B}$, and two direction vectors, ${\bf{vl}},{\bf{vr}}$. It represents the velocity set that might incur collisions for a robot within a period. To achieve collision avoidance from obstacles, robot $A$ should select a velocity outside the VO area, that is ${{{\bf{v'}}}_A} \notin VO_B^A({{\bf{v}}_B})$. Similarly, the VO area of the static line obstacle is constructed using the same definition, which is shown in Fig.~\ref{rvo} (b). In addition, for the situation that the static obstacle blocks the path from the current position of a robot to its goal, a global navigation strategy is required to generate a rational desired velocity.

As VO based methods have been widely used for avoiding dynamic obstacles, RVO is more suitable for a group of robots to avoid each other actively. As illustrated in Fig.~\ref{rvo} (c), RVO is extended from the VO concept geometrically and can be described using the following formula: 
\begin{equation}
RVO_B^A({{\bf{v}}_B},{{\bf{v}}_A}) = \{ {{{\bf{v'}}}_A}\left| {2{{{\bf{v'}}}_A} - {{\bf{v}}_A} \in VO_B^A({{\bf{v}}_B})} \right.\}.  
\end{equation}
Actually, a RVO area can be obtained by translating from a VO area whose apex is moved from ${{{\bf{v}}_B}}$ to $\frac{{{{\bf{v}}_A} + {{\bf{v}}_B}}}{2}$. Thus, both VO and RVO can be represented by a six-dimensional vector $\mathbf{c}=\left[\mathbf{v}_{p}, \mathbf{v} \mathbf{l}, \mathbf{v} \mathbf{r}\right] \in \mathbb{R}^{6}$, where ${{\bf{v}}_p} = [{v_x},{v_y}]$ denotes the coordinates of the apex, ${\bf{vl}} = [v{l_x},v{l_y}]$ and ${\bf{vr}} = [v{r_x},v{r_y}]$ describe the direction of left and right rays respectively. Within a robot group, each robot selects a velocity outside such a joint VO and RVO areas to achieve the collision avoidance task cooperatively.

\subsection{Problem Statement}

The multi-robot collision avoidance problem can be defined as an optimization problem that finds a series of optimal velocities to minimize the travel time under the collision avoidance constraints. For a group of $n$ differential drive robots navigating in the shared workspace, each robot $i$ with radius $R_i$, state ${\bf{p}}_t^i = [p_{xt}^i,p_{yt}^i]$, and velocity ${\bf{v}}_t^i = [v_{xt}^i,v_{yt}^i]$ can sense $m$ surrounding robots $j$ with a limited range $d_l$ at time $t$. The robot's observation has two parts, the proprioceptive measurement ${{\bf{o}}_{\text{self}}}$ of the state of the ego-robot and the exteroceptive measurement ${{\bf{o}}_{\text{sur}}}$ of the surrounding environment. The proprioceptive observation includes the robot's current velocity, orientation, its desired velocity, and its virtual radius for collision avoidance, \textit{i.e.}, ${{\bf{o}}_{\text{self}}} = [{{\bf{v}}_t},\emph{ori},{\bf{v}}_t^{\text{des}},{R_c}]$. The exteroceptive observation contains the RVO vectors ${\bf{c}}$, as well as the relative distance $d$ and expected collision time ${t_e}$ between the ego-robot and its $j$-th neighboring robot or obstacle under the current velocity. While the expected collision time will be an infinite value when there is no collision possible under the current situation, which will influence the subsequent calculations. Thus, we use the reciprocal value ${r_e}$ to replace that, i.e., ${r_e} = 1/({t_e} + 0.2)$, where $0.2$ is a constant value. Hence, the exteroceptive observation should be: ${\bf{o}}_{{\rm{sur}}}^j = [{{\bf{c}}_j},{d_j},{r_e}_j]$, where $j = 0,1,\cdots, m$. Given the proprioceptive/exteroceptive observations as the input, the policy neural network ${\pi _\theta }$ parameterized by $\theta $ will output the actions ${\mathbf{a}_t}$, which are the robot's velocity increment in each time step, \textit{i.e.}, ${\Delta\mathbf{v}_t} = [\Delta {v_{xt}},\Delta {v_{yt}}]$. The computed velocity increments are optimal if the resulting velocity in the next time step is not only collision-free but also has a minimal difference to the robot's desired velocity. In other words, we need to solve the following constrained optimization problem at each time step $\Delta t$ for each robot:

\begin{equation}
    \begin{array}{ll}
        \underset{\pi_{\theta}}{\arg \min } & \left\|\mathbf{v}_{t}-\mathbf{v}_{t}^{\operatorname{des}}\right\|, \\
        \text { s.t. } & \Delta \mathbf{v}_{t} \sim \pi_{\theta}\left(\mathbf{a}_{t} \mid \mathbf{o}_{\text {self }}, \mathbf{o}_{\text {sur }}\right), \\
        & \mathbf{v}_{t}=\mathbf{v}_{t-1}+\mu \cdot \Delta \mathbf{v}_{t}, \\
        & \mathbf{p}_{t}=\mathbf{p}_{t-1}+\Delta t \cdot \mathbf{v}_{t}, \\
        & d_{j}=\left\|\mathbf{p}_{t}-\mathbf{p}_{j t}\right\| ,\\
        & \forall j \in[1, m], d_{j}>R_{c}+R_{j c},
        \end{array}
\end{equation}
where $\left\|  \cdot  \right\|$ is the Euclidean norm operation for a vector, $\mu$ is the hyperparameter adjusting the range of the velocity increment, and ${d_j}$ is the distance between a robot and its $j$-th neighbor with position $\mathbf{p}_{j t}$ and collision radius $R_{j c}$. All the robots share the same navigation policy $\pi_{\theta}$ and find the optimal velocity independently.

\section{Reinforcement Learning Framework}\label{section 4}

An actor-critic reinforcement learning framework consists of observation space, action space, reward function, policy actor, and policy critic. The policy actor in the form of a deep neural network maps the observed information into an action to control the robot. The policy critic in the form of a deep neural network utilizes the reward function to evaluate each action made by the actor.

\subsection{Observation Space and Action Space}
The observation space contains the proprioceptive and exteroceptive measurements, as mentioned in Section~\ref{section 3}, ${\bf{o}} = [{{\bf{o}}_{\text{self}}},{\bf{o}}_{\text{sur}}^j], j = 0,\cdots, m$. The action space is the velocity increment within the $x-y$ plane, ${\bf{a}} = [\Delta {v_x},\Delta {v_y}]$. Thus, the control vector at the next time step should be ${{\bf{v}}_{t + 1}} = {{\bf{v}}_t} + \mu  \cdot {\bf{a}}$. Specifically, the velocity is clipped within a range between a maximum and a minimum value, ${{\bf{v}}_t} \in [{{\bf{v}}_{\min }},{{\bf{v}}_{\max }}]$. However, all the robots are non-holonomic and controlled by the transitional and rotational velocity, ${\bf{v}}_t^c = [{v_t},{\omega _t}]$. Thus, we convert the orthogonal velocity ${{\bf{v}}_t} = [{{\bf{v}}_{xt}},{{\bf{v}}_{yt}}]$ to ${\bf{v}}_t^c$ as follows:
\begin{equation}
    \left\{ {\begin{array}{*{20}{l}}
        {{v_t} = \left\| {{{\bf{v}}_t}} \right\| \cdot \cos (\varsigma )}\\
        {{\omega _t} =  - \varsigma /\tau }
        \end{array}} \right.,
\end{equation}
where $\varsigma $ is the radian difference between the orientations of the robot and velocity ${{{\bf{v}}_t}}$, and $\tau $ is the guaranteed time to adjust the rotation rate. Thus, the transitional velocity is determined by the speed component along with the robot orientation. And the rotational velocity is to make the robot orientation consistent with the direction of orthogonal velocity. 

\subsection{ Neural Network Architecture }

\begin{figure}[tp]
    \centering
    \includegraphics[width=0.40\textwidth, clip]{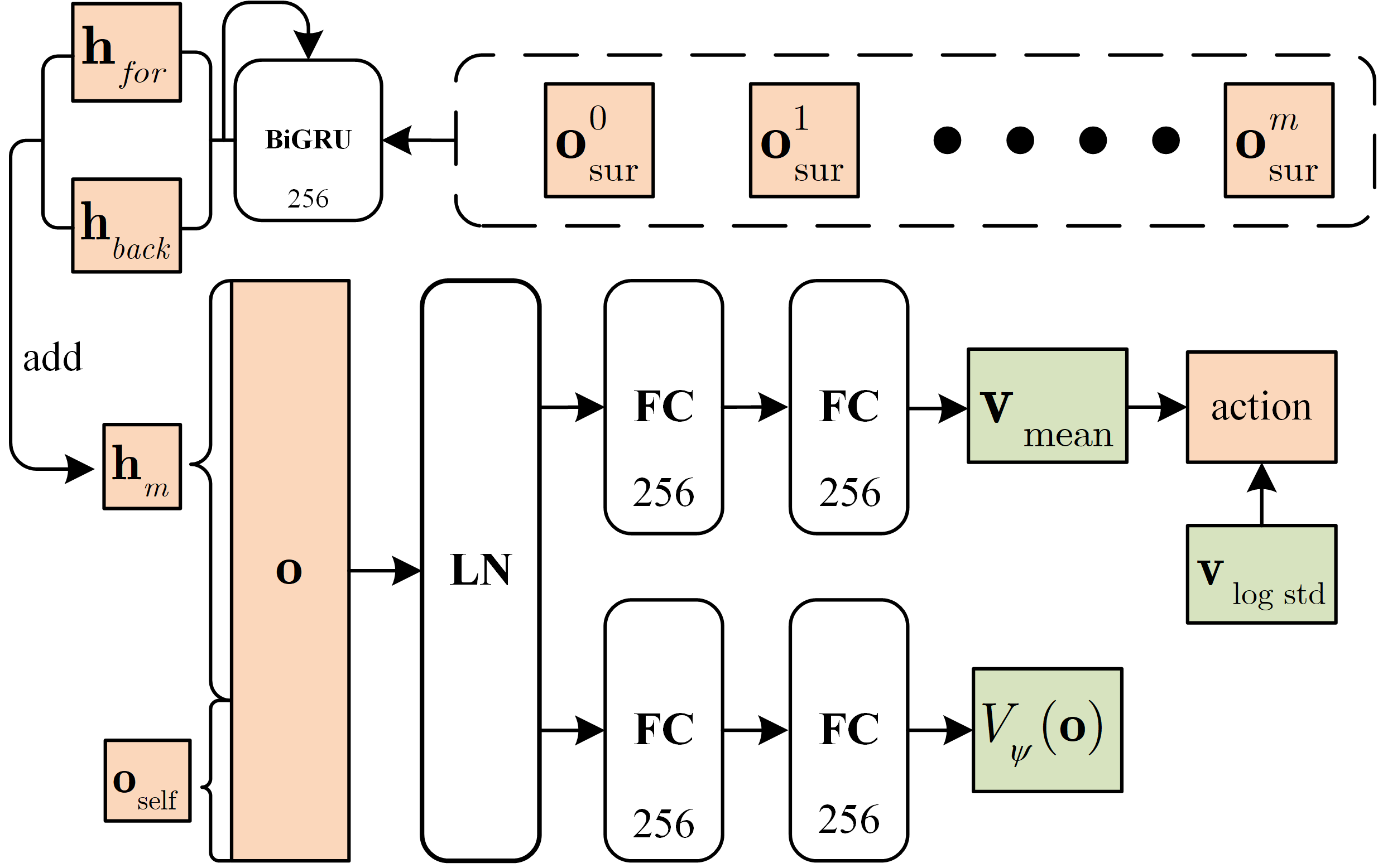}
    \caption{An illustration of the proposed BiGRU based neural network architecture of the navigation policy actor and critic.}
    \label{dnn}
\end{figure}

The architecture of the deep neural network is illustrated in Fig.~\ref{dnn}. For the variable-length input sequences, existing work uses the LSTM to produce a fixed-length representation, which processes inputs in a strict order~\cite{everett2018motion}. By contrast, we use the bidirectional GRU module, which consists of two GRUs to process the inputs with both forward and backward directions, respectively. Such a scheme helps the network more accurately to find the underlying relationship among input vectors better depending on the limited information, resulting in a better presentation of the input sequences. At each step, the exteroceptive measurements ${\bf{o}}_{\text{sur}}^j$ in terms of a series of vectors are fed into the BiGRU in the first ascending order of ${r_e}$ and the second descending order of the distance $d$. There are two final hidden states, ${{\bf{h}}_{for}}$ and ${{\bf{h}}_{back}}$, from the forward and backward GRU, respectively. They are added as one 256-D fixed-size vector ${{\bf{h}}_m}$. This output is concatenated with the proprioceptive measurement ${{\bf{o}}_{\text{self}}}$ as the integrated fixed-length observation ${\bf{o}}$. To achieve a normalized representation to speed up the training process, where all parts have a common scale, we use the Layer Normalization method~\cite{ba2016layer} to process fixed-length observation. The output is used as the input data of two neural networks. One is for the policy actor ${{\pi _\theta }}$, which consists of two hidden fully connected (FC) layers with 256 rectifier linear units. The output of the final layer is a two-dimensional vector, which is the mean of velocity ${\bf{v}}{}_{\text{mean}}$, activated by a hyperbolic tangent activation function (Tanh). The action is sampled from a Gaussian distribution represented by its mean, ${\bf{v}}{}_{\text{mean}}$, and log standard deviation ${\bf{v}}{}_{\text{log std}}$, where ${\bf{v}}{}_{\text{log std}}$ is updated independently. Another neural network is for the policy critic ${V_\psi }({\bf{o}})$, which is also composed of two hidden fully-connected layers with 256 rectifier linear units. The output of its final layer activated by an identity function is a scalar value used to evaluate each state-action pair. 

\subsection{Reward Function}
In the reinforcement learning framework, the design of the reward function is vital to policy performance. The common distance based reward function, such as the penalties for the increasing distance to obstacles or the awards for the decreasing distance to the goal~\cite{long2018towards, everett2018motion}, is improper for our RVO observations, which do not include the information of positions and velocities directly. Thus, a novel reward function based on RVO areas is developed, which helps each robot to learn a reciprocal collision avoidance behavior. In addition, based on our previous experiences, the expected collision time performs better than the distance-to-obstacle in terms of representing the collision risk, as the expected collision time takes both the distance-to-obstacle and relative velocity between the robot and the obstacle into account. Based on this concept, the RVO reward function $r_{\text{rvo}}^t$ at time $t$ is designed as follows, which represents the quality of the selected velocity ${{\bf{v}}_t}$ judged by the joint RVO area:
\begin{equation}
    r_{rvo}^t = \left\{ {\begin{array}{*{20}{l}}
        {a - b*dif{f_v}}&{{\rm{ if\ }}{{\bf{v}}_t} \notin RVO\ { \rm{ or\ }}\xi  > 5}\\
        {c - d*{{(\xi  + f)}^{ - 1}}}&{{\rm{ if\ }}{{\bf{v}}_t} \in RVO\ {\rm{ and\ }}\xi  > 0.1}\\
        { - e*{{(\xi  + f)}^{ - 1}}}&{{\rm{ if\ }}\xi  \le 0.1}
        \end{array}} \right.,
    \label{rvo_reward}
\end{equation}
where $\diffv_v$ is the distance between the selected velocity and desired velocity, $\diffv_v = \left\| {{{\bf{v}}_t} - {\bf{v}}_t^{\text{des}}} \right\|$, and $\xi$ is the expected minimum time that the robot has a collision with an obstacle under the current velocity. Whether the current velocity ${{\bf{v}}_t}$ is within the RVO area is judged by a six-dimensional vector ${\bf{c}} = [{{\bf{v}}_p},{\bf{vl}},{\bf{vr}}]$ as follows:

\begin{small}
\begin{equation}
    \left\{ {\begin{array}{*{20}{c}}
        {{{\bf{v}}_t} \in RVO:{\rm{ = (}}{{\bf{v}}_t} - {{\bf{v}}_p}) \times {\bf{vl}} \ge 0 \wedge {\rm{(}}{{\bf{v}}_t} - {{\bf{v}}_p}) \times {\bf{vr}} \le 0}\\
        {{{\bf{v}}_t} \notin RVO:{\rm{ = (}}{{\bf{v}}_t} - {{\bf{v}}_p}) \times {\bf{vl}} < 0 \vee {\rm{(}}{{\bf{v}}_t} - {{\bf{v}}_p}) \times {\bf{vr}} > 0}
        \end{array}} \right.,
\end{equation} 
\end{small}  
where $\times $ is cross product.

The characters $a, b, c, d, e, f$ are all constant values and tunable to adjust the policy performance. $a$ and $c$ are the basic awards to encourage the behavior when the current velocity is close to the desired velocity, or the current collision risk is low, which tend to be positive values. $b$ is the coefficient to adjust the weight of $dif{f_v}$, which represents the motion of moving with the desired velocity. While $d, e$ are the coefficients to adjust the weight of ${{(\xi  + f)}^{ - 1}}$, which denotes the motion of keeping longer expected collision time. The range of ${{{(\xi  + f)}^{ - 1}}}$ is  $[0,{f^{ - 1}}]$. It is the substitute of $\xi$, which is inappropriate for policy training directly because of the range $[0, + \infty ]$. To summarize, for the crowded scenarios, bigger values of $d$ and $e$ are recommended to guarantee the collision avoidance ability. The values for our policy training in this paper are $a = 0.3,b = 1.0,c = 0.3,d = 1.2,e = 3.6,f=0.2$. The main idea is that when the velocity is in the joint RVO area, the robot has to pay attention to the potential collision judged by the expected collision time. The constant values of the reward function are all tunable parameters, which can be adjusted to improve the training performance. 

\section{Policy Training}\label{section 5}

\begin{algorithm}[t]
    \caption{Policy Training Algorithm}
    \label{pt}
    Initialize the policy actor ${\pi _\theta }$ and critic ${V_\psi }$\;
    \For{epoch $\leftarrow1, 2,\cdots$}
    {
	    \For(\tcp*[h]{collect data}){robot $i \leftarrow 1$ \KwTo $n$}
	    {
	        Run policy ${\pi _{\theta {\rm{old}}}}$ for $T$ steps\;
	        Collect data $\{ {\bf{o}}_t^i,{\bf{a}}_t^i,r_t^i\}$ for $T$ steps\;
	        Compute GAE: $\hat A_1^i,\cdots, \hat A_T^i$\;
	        Add data into \emph{buffer} $i$\;
	    }
	    \For(\tcp*[h]{update policy}){buffer $j \leftarrow 1$ \KwTo $n$}
	    {
	        \For{$k \leftarrow 1$ \KwTo $K_{iter}$}
	        {
	            Optimize $L^{CLIP}(\theta)$ w.r.t. $\theta$ with ${l_a}$\;
	            \lIf{KL-divergence $> \overline{KL}$}{\bf break}
	            ${\theta _{old}} \leftarrow \theta$\;
	        }
	        \For{$h \leftarrow  1$ \KwTo $H_{iter}$}
	        {
	            Optimize ${L^V}(\psi )$ w.r.t. $\psi $ with ${l_v}$\;
	            ${\psi _{old}} \leftarrow \psi $\;
	        }
	    }
   }
\end{algorithm}

PPO is a popular optimization method for DRL algorithms, which is advantageous for training the policy associated with continuous action spaces. In this work, we utilize the PPO algorithm to train and update the multi-robot collision avoidance policy successively. The detail of the training algorithm is summarized in Algorithm~\ref{pt}. First, this algorithm begins with the initialization of the neural network parameters, including the policy actor ${\pi _\theta }$ and critic ${V_\psi }$. Then, during the training loop, each robot runs in the environment with the navigation policy ${\pi _\theta }$ for $T$ timesteps. The data of observation, reward, and action of each robot at each time step $t$ is collected. Generalized advantage estimator (GAE) is used to estimate the advantage of the action of robot $i$ using the reward function $r_t^i$ and value function ${V_\psi }$. The related observation, action, and reward data are stored in the buffer for the following policy update. The buffered data are used to construct the clip surrogate objective $L^{CLIP}(\theta )$ and loss of value function ${L^V}(\psi )$~\cite{schulman2017proximal}, which are optimized with the Adam optimizer and learning rate ${l_a}$, ${l_v}$ respectively. The learning rates determine the change rate of the policy model within each update step. The Kullback-Leiber (KL)-divergence measures the difference between two probability distributions of the previous and current policies. The update loop will stop when the KL-divergence value is over a threshold (denoted by $\overline{KL}$), which means that the model changes too fast, and hence such an update process is unsuccessful. 

\section{ Experiments and Results}\label{section 6}

\subsection{Simulation Setup}

\begin{figure}[hp]
    \centering
    \includegraphics[width=0.48\textwidth, clip]{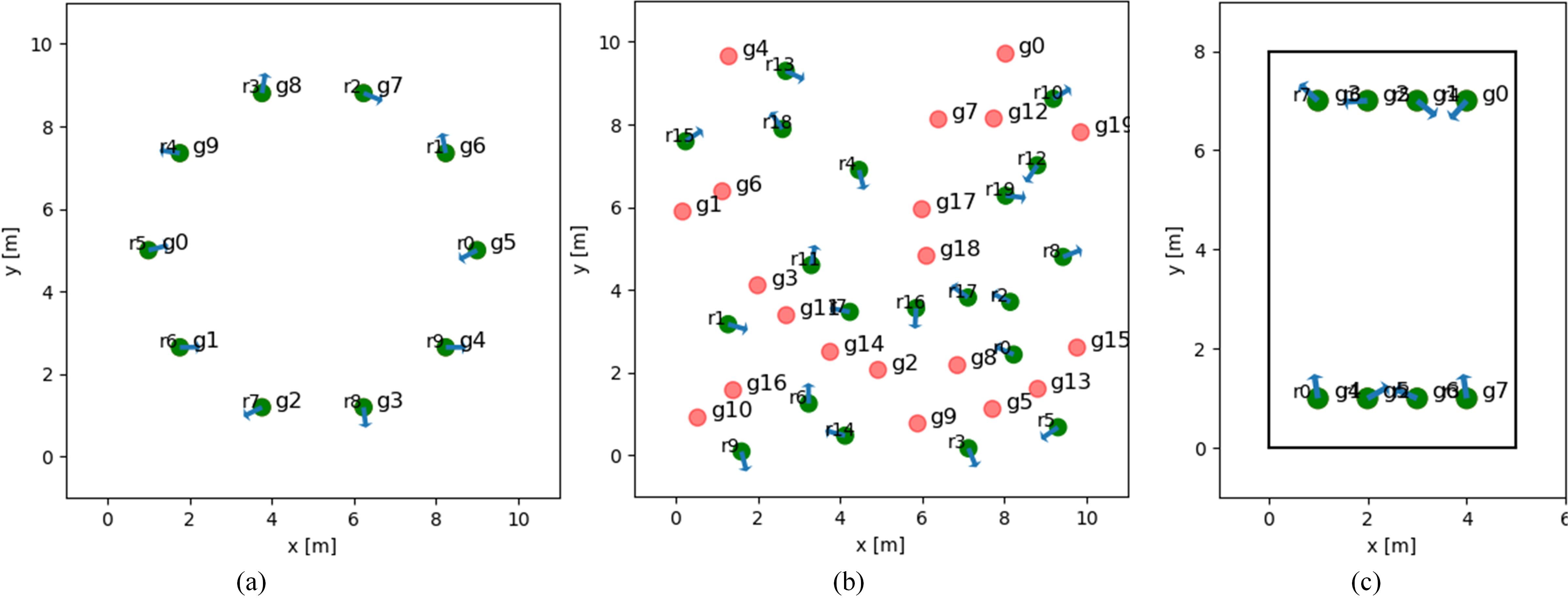}
    \caption{The simulated scenarios for multi-robot navigation: (a) circle scenario, (b) random scenario, (c) corridor scenario.}
    \label{scen}
    \vspace{-0.5cm}
  \end{figure}

In this work, the navigation policy RL-RVO is trained through the simulated scenario and implemented with Pytorch (Python 3.8). The simulated scenarios are developed and plotted via the OpenAI Gym interface~\cite{brockman2016openai}, which is a popular toolkit for developing reinforcement learning algorithms. We use several disc-shaped robots to train the collision avoidance policy through the circle scenario only. In the circle scenario, all robots with random orientations are uniformly arranged along with a circle shape. And the goal position is on the opposite side, which leads to a rich interaction for the robots, as illustrated in Fig.~\ref{scen} (a). Specifically, the sensing range $d_l$ and the maximum input number of neighbors $\kappa$ of each robot are limited to be $4m$ and 5, respectively. All robots share the same policy and collect the observations for the parameter update by the PPO algorithm after each epoch. The environment will be reset when there are collisions or over the maximum episode length. To speed up the training convergence, we use two stages to perform the training process. First, the policy is trained in a $9m \times 9m$ circle scenario with a few robots ($2$ or $4$) for $e_1$ epochs to achieve basic functionality, such as moving toward the goal position. Then, the policy continues to be trained with $10$ robots in the same circle scenario for about $e_2$ epochs. The training process will be early stopped depending on the success rate and step cost. The values of the hyperparameters for the training process are chosen, as listed in Table~\ref{parameter}. Typically, the training process is performed via a computer with CPU i7-9700 and GPU Nvidia GTX 1080 for about 10 hours. 

\begin{table}[]
    \centering
    \caption{The hyperparameters of the training process}
    \label{parameter}
    \begin{tabular}
    {cc|cc|cc|cc}
    \hline
    Para. & Val. & Para. & Val. & Para. & Val. & Para. & Val. \\ \hline
    ${d_l} $                    & 4     & ${l_a}$       &  4e-6  & ${K_{{iter}}} $               & 50  &   $\kappa $    &  5  \\
    $ \tau$                     & 0.2   & ${l_v}$       &  5e-5  & $\overline{KL}$                   & 0.01   & $\mu $        & 1.0  \\
    ${R_c}$                     & 0.3   & ${e_1}$       &  200 & ${H_{{iter}}}$                & 50    &  $T$          & 450 \\
    $R_{j c}$                    & 0.3   & ${e_2}$       &  1000 & ${{\bf{v}}_{\min }}$        & -1.5    &  ${{\bf{v}}_{\max }}$          & 1.5 \\\hline 
    \end{tabular}
\end{table}

\subsection{Results and Discussions}

\subsubsection{Metrics}
Three metrics are utilized to evaluate the policy performance, including success rate, travel time, and average speed. The success rate is a ratio of the successful cases without any collision or being stuck somewhere during the navigation, which describes the policy's ability of collision avoidance. The travel time refers to the amount of time, represented by the iteration step in the simulation when all robots arrive at the goal positions, which reflects the policy's efficiency. The average speed of the navigation process of the whole robot team measures the policy's performance on effective velocity selection. In addition, the average computational cost of a control action for one robot is also compared.

\subsubsection{Simulated experiments}

\begin{figure}[hp]  
    \begin{subfigure}{0.23\textwidth}
    \hfill\subcaptionbox{RL-RVO}{\includegraphics[width=0.7\linewidth]{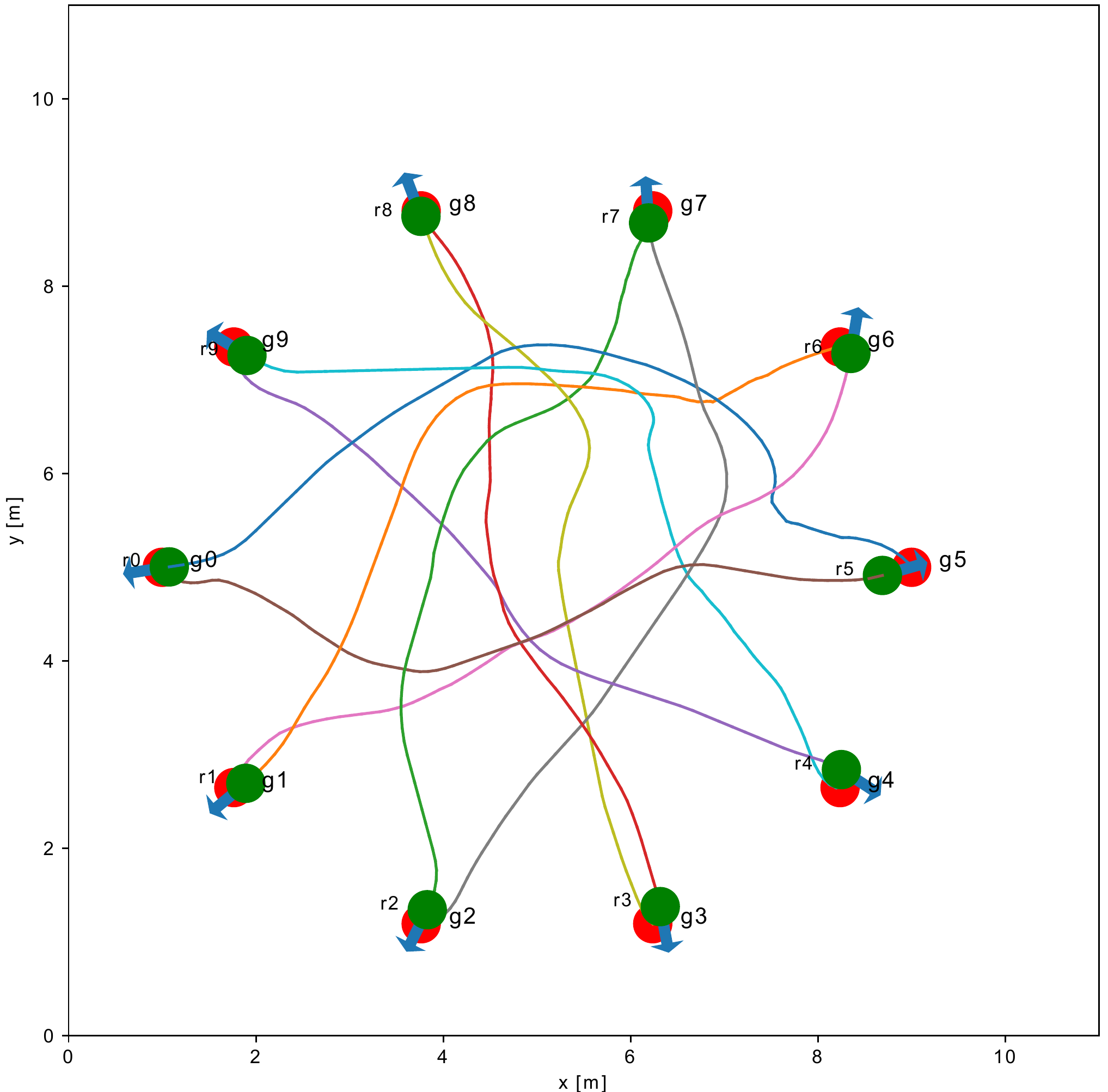} }
    \end{subfigure}
    \hfill
    \begin{subfigure}{.23\textwidth}
    \subcaptionbox{SARL}{\includegraphics[width=0.7\linewidth]{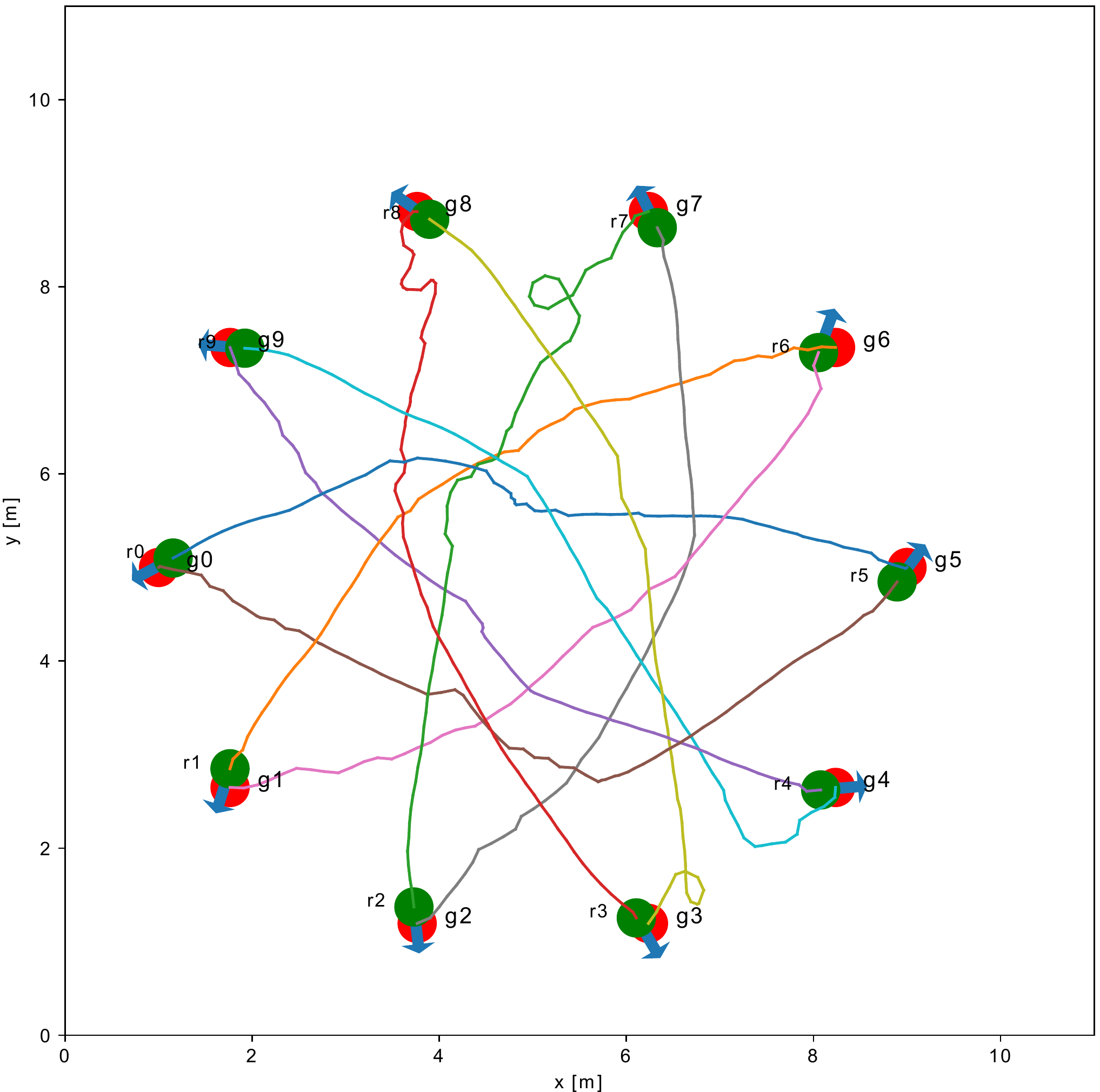}}
    \end{subfigure}
  
    \begin{subfigure}{.23\textwidth}
    \hfill\subcaptionbox{GA3C-CADRL}{\includegraphics[width=0.7\linewidth]{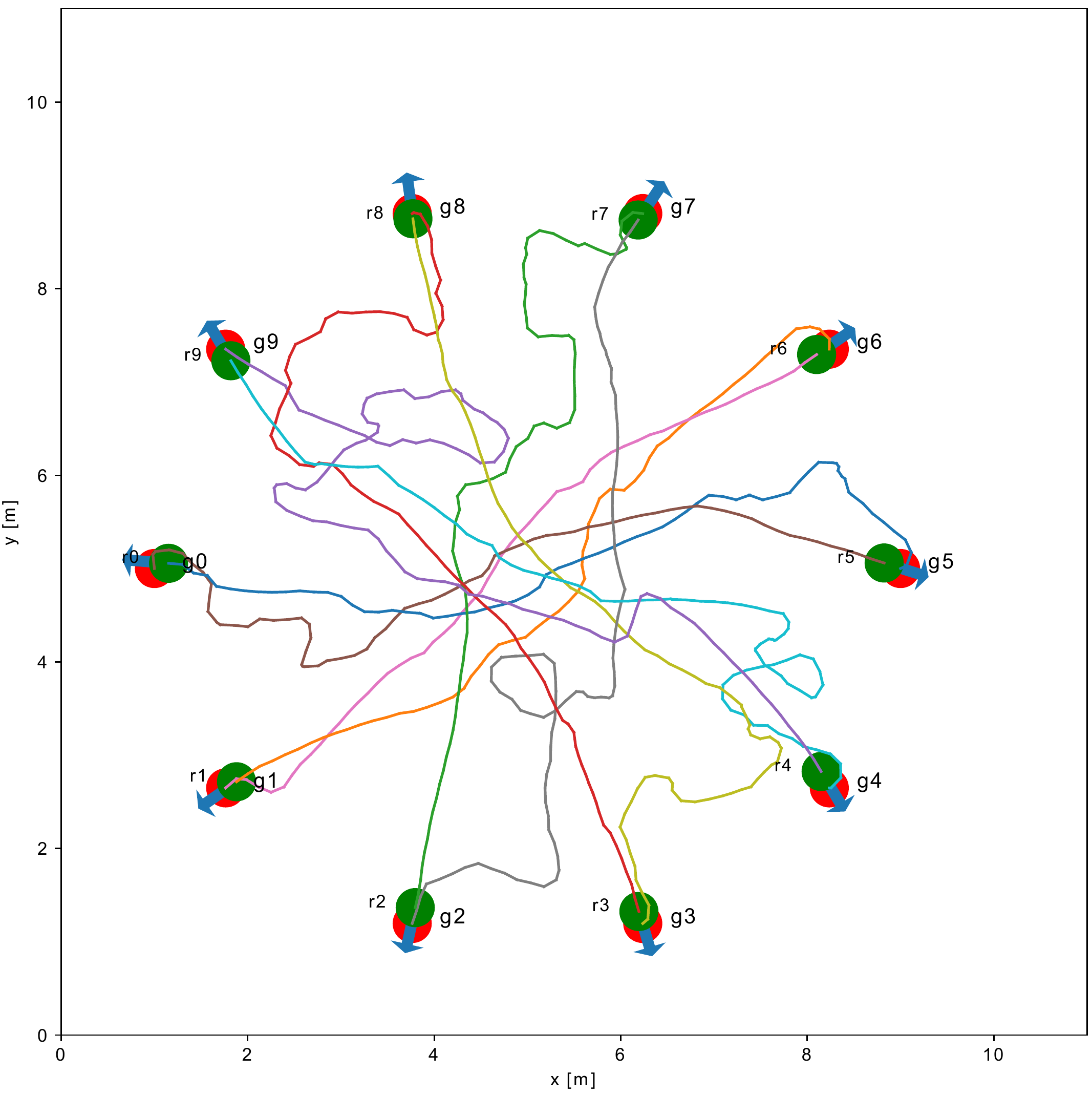} }    
    \end{subfigure}
    \hfill
    \begin{subfigure}{.23\textwidth}
    \subcaptionbox{NH-ORCA}{\includegraphics[width=0.7\linewidth]{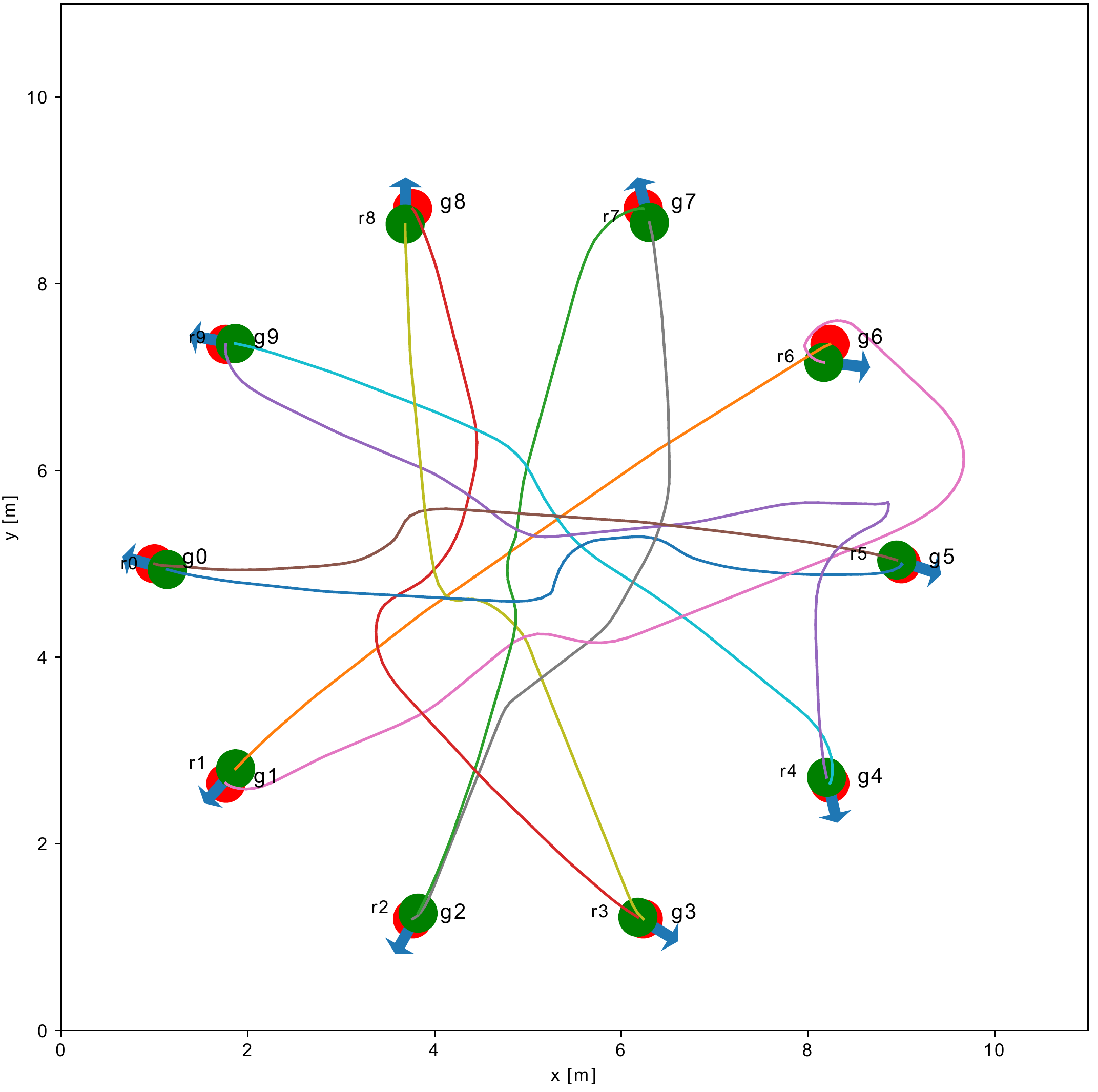} }    
    \end{subfigure}
  
    \caption{An illustration of the trajectories generated by four approaches respectively in the circle scenario.}
    \label{traj}
  \end{figure}

\begin{figure}[hp]
    \centering
    \includegraphics[width=0.20\textwidth, clip]{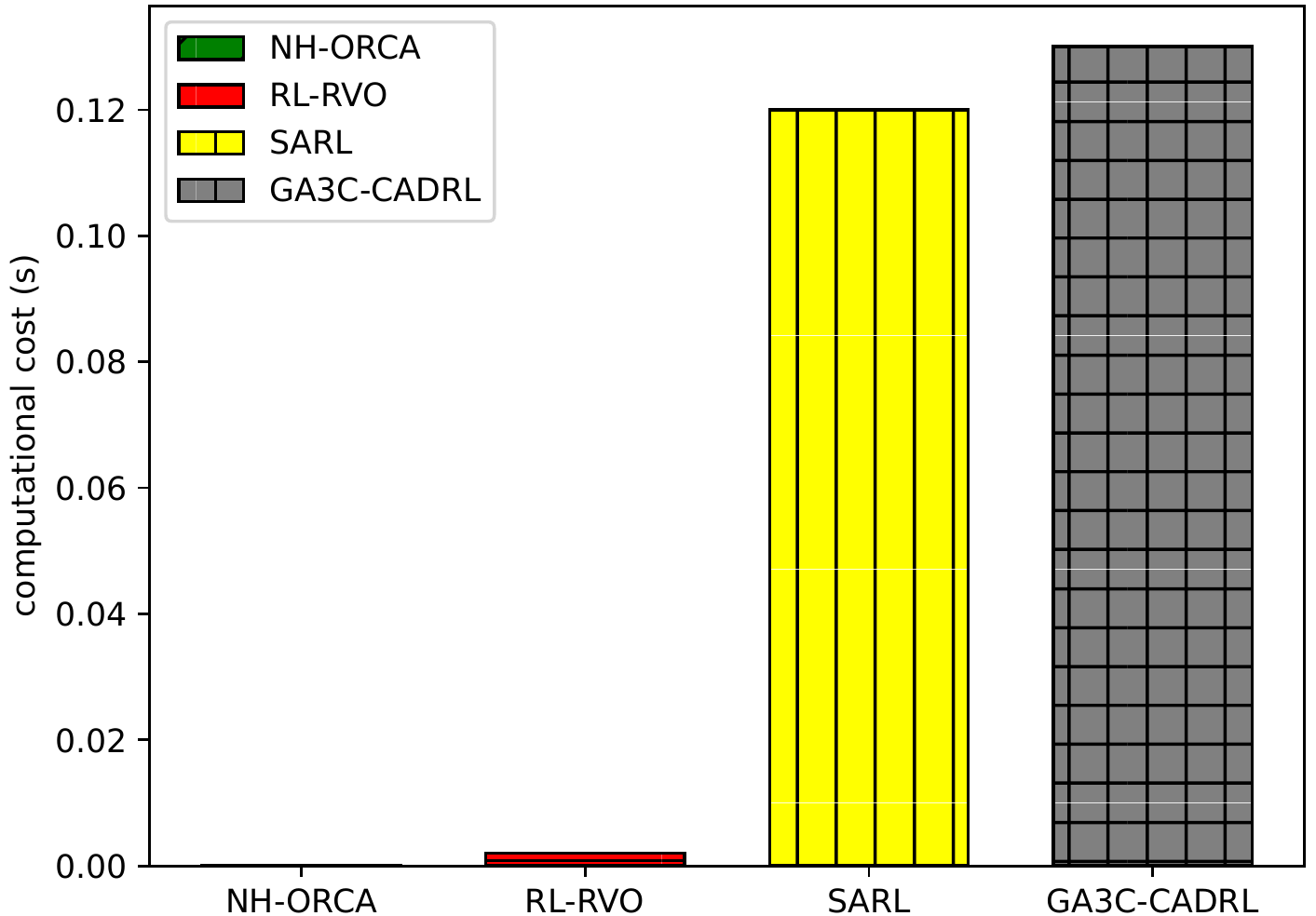}
    \caption{A comparison of the average computational cost to achieve single control action for each robot.}
    \label{cost}
    \vspace{-0.3cm}
\end{figure}

To validate the collision avoidance performances of our policy, we compare it with SARL~\cite{chen2019crowd}, GA3C-CADRL~\cite{everett2018motion}, and NH-ORCA~\cite{alonso2013optimal} in both circle and random scenarios. The random scenarios are used to test the generalization ability of a navigation policy trained from a specific scenario, where both initial and target positions are generated randomly with a minimum interval ($1m$) and change randomly in each test episode, as shown in Fig.~\ref{scen} (a) and (b). To test the ability of collision avoidance under limited information, the sensing range is set to be ${d_l}{\rm{ = }}4m$. All policies are performed for $100$ episodes with various numbers of robots (from $6$ to $20$). The average results with standard deviations (std) of $100$ cases in $9m \times 9m$ circle scenarios and $10m \times 10m$ random scenarios are listed in Table~\ref{results}. The trajectories of four approaches are compared visually in Fig.~\ref{traj}. Finally, the average computational cost to achieve single control action for each robot is shown as a bar figure in Fig.~\ref{cost}.


\vspace{-0.5cm}

\begin{figure}[tp]
    \begin{subfigure}[b]{0.45\textwidth}
        \includegraphics[width=0.33\linewidth]{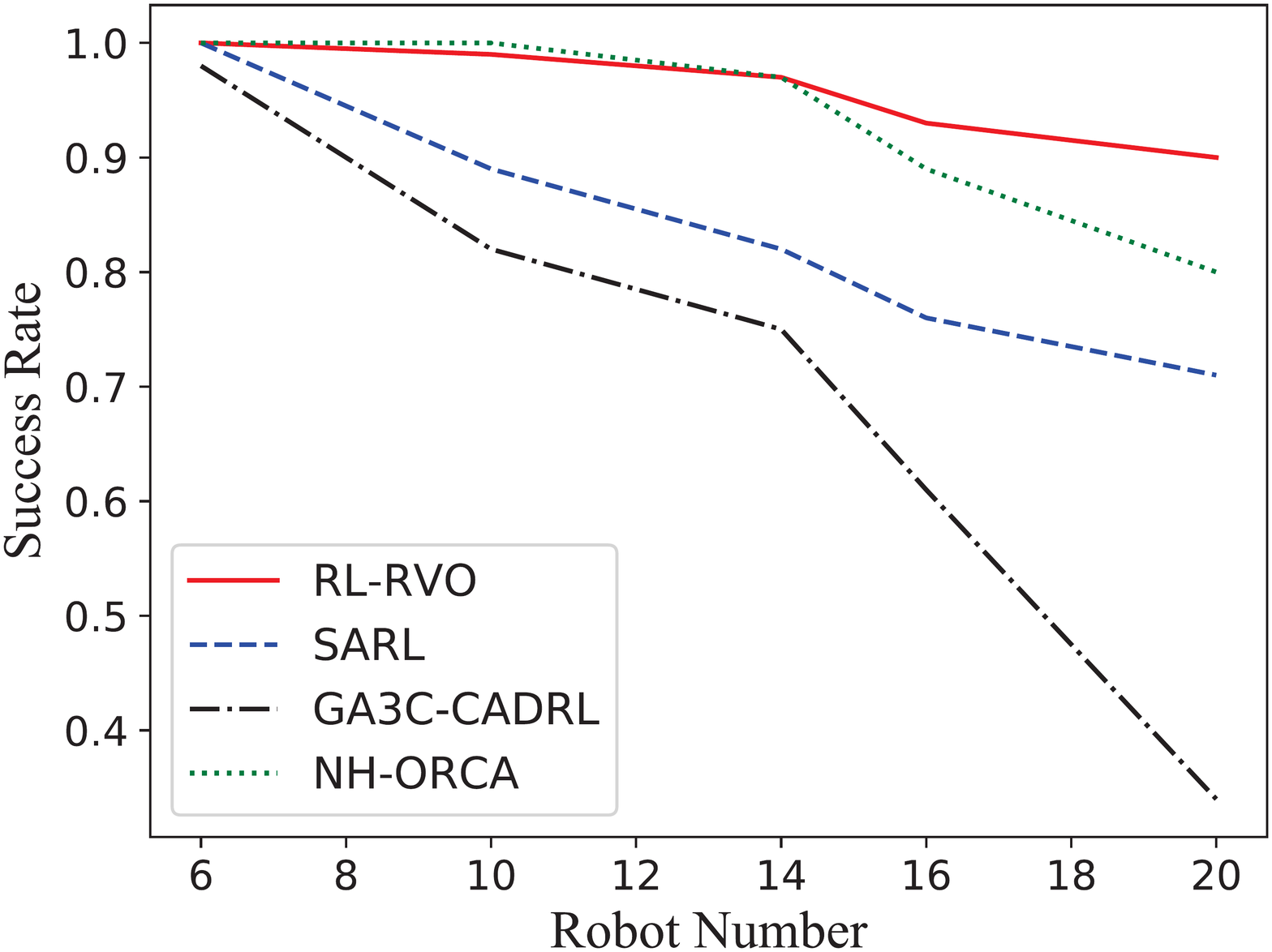}\hfill
        \includegraphics[width=0.33\linewidth]{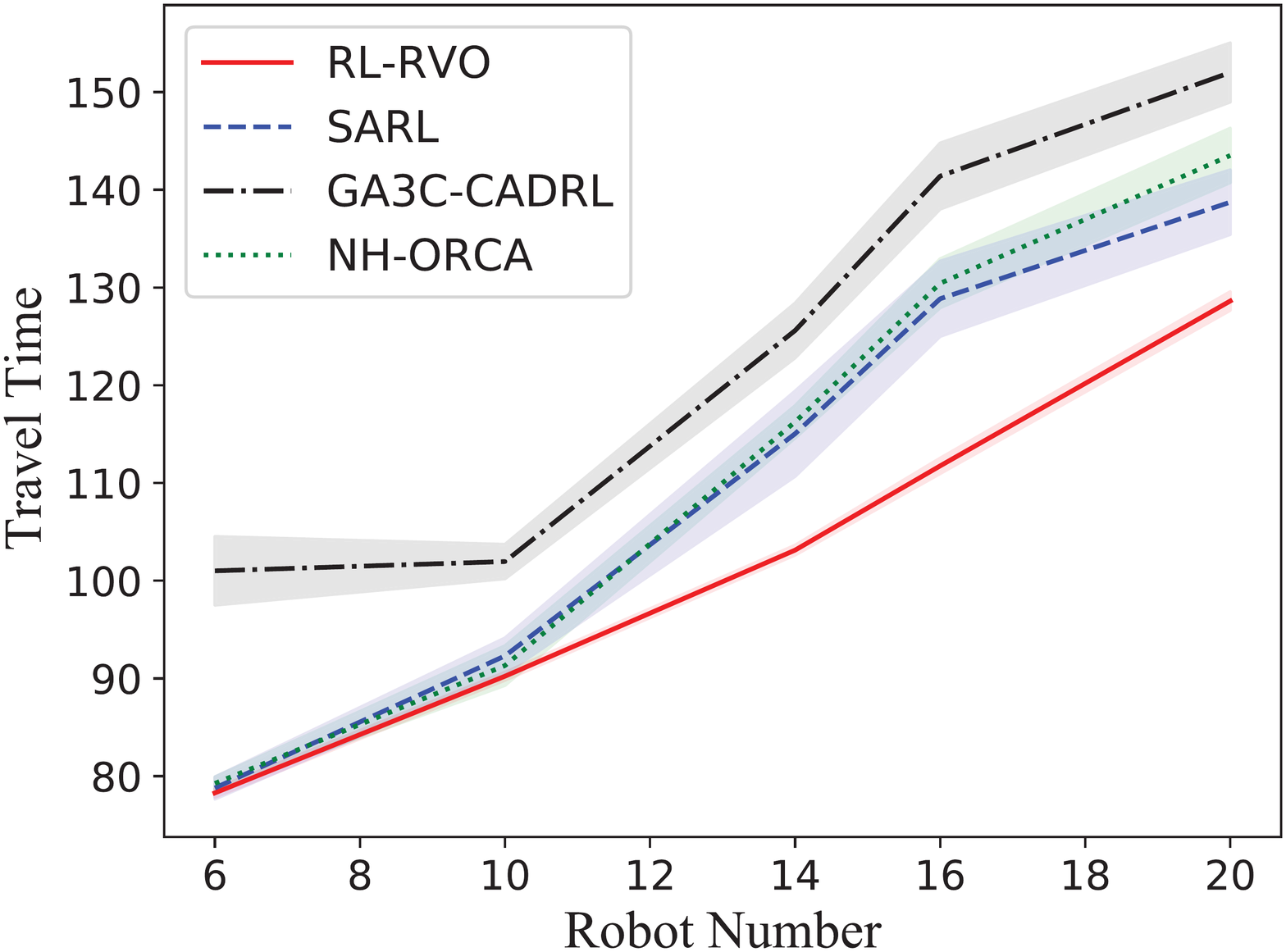}\hfill
        \includegraphics[width=0.33\linewidth]{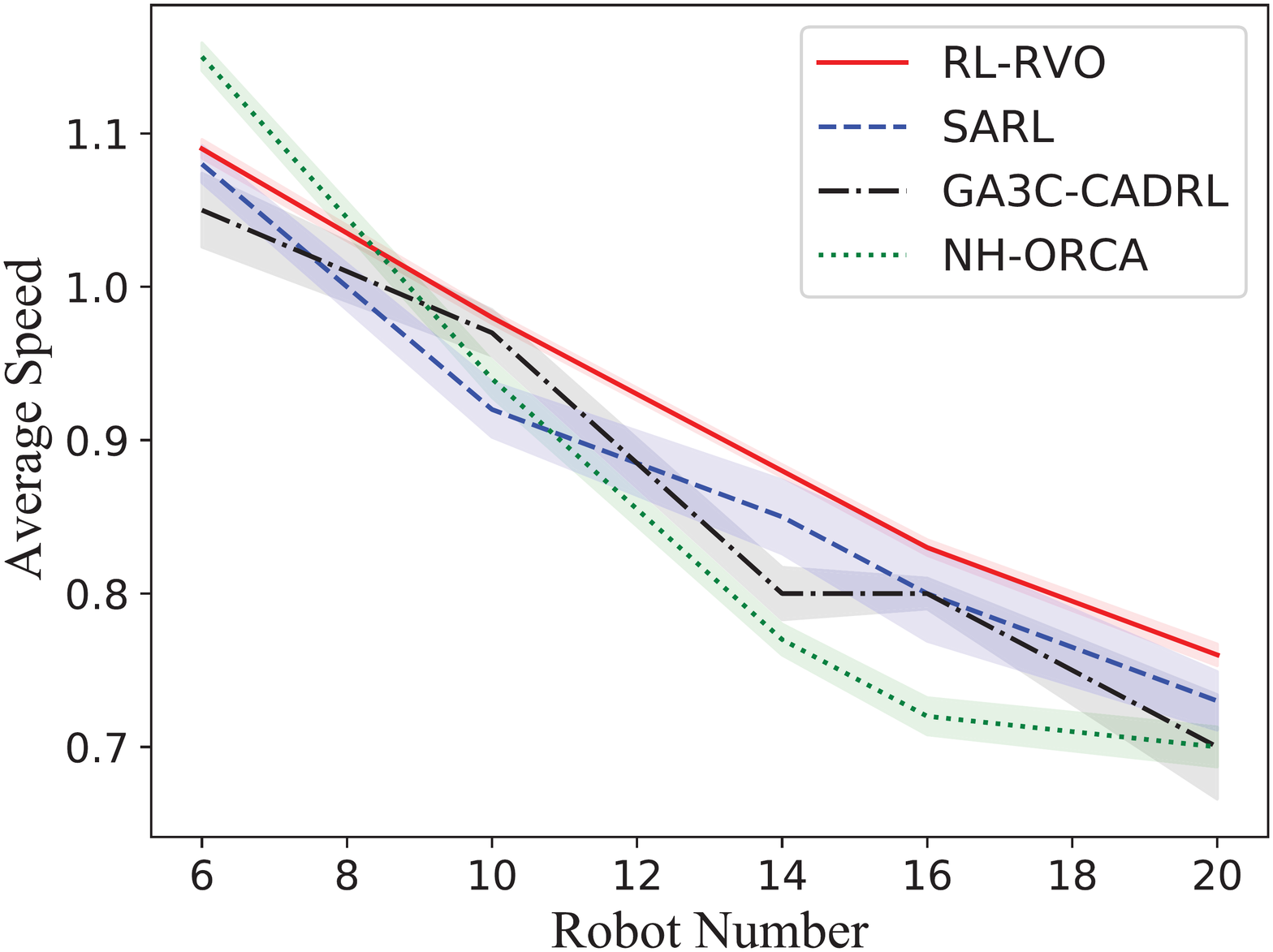}\hfill
        \subcaption{The results in the circle scenario.}
    \end{subfigure}
    \begin{subfigure}[b]{0.45\textwidth}
        \includegraphics[width=0.33\linewidth]{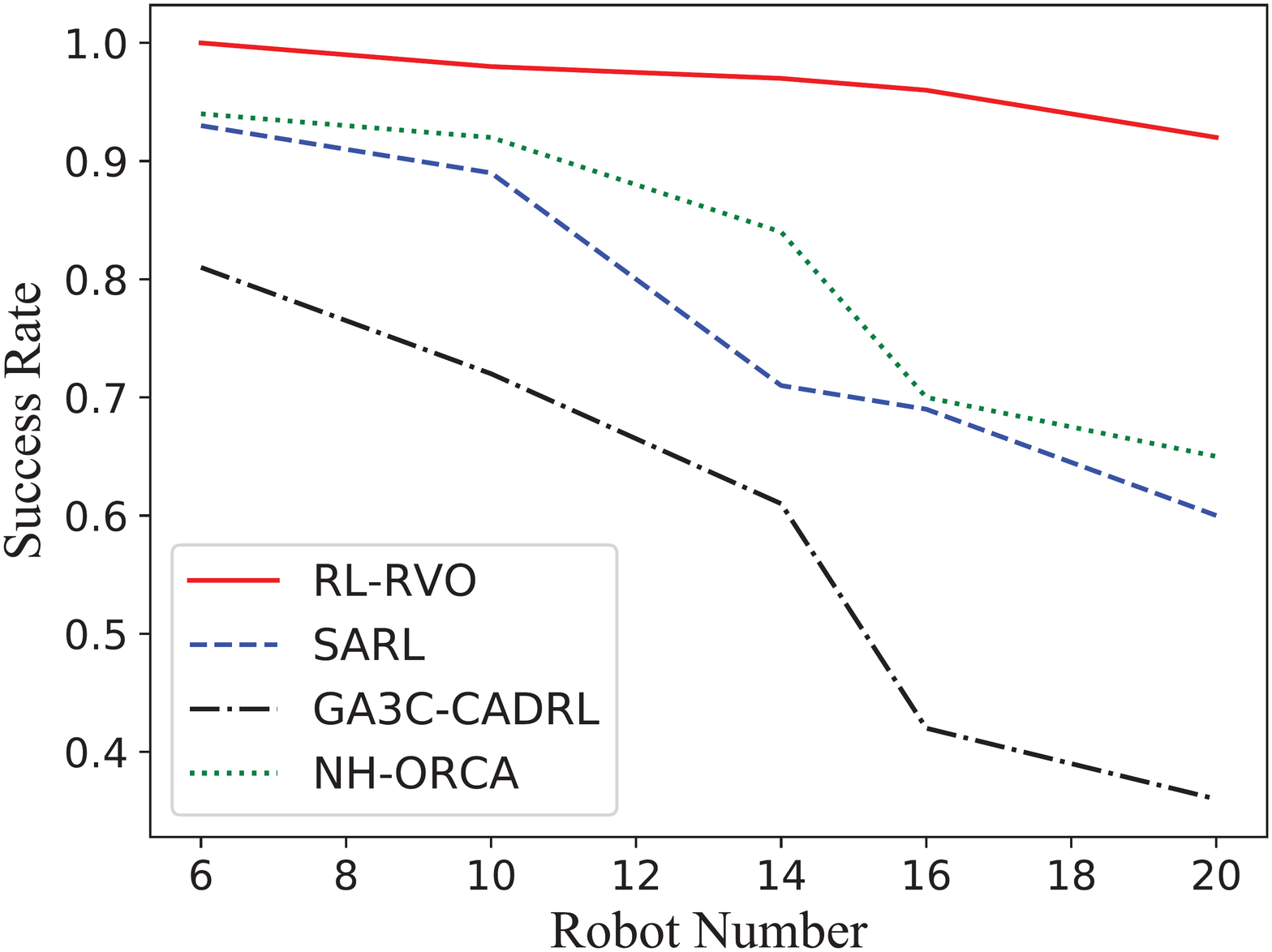}\hfill
        \includegraphics[width=0.33\linewidth]{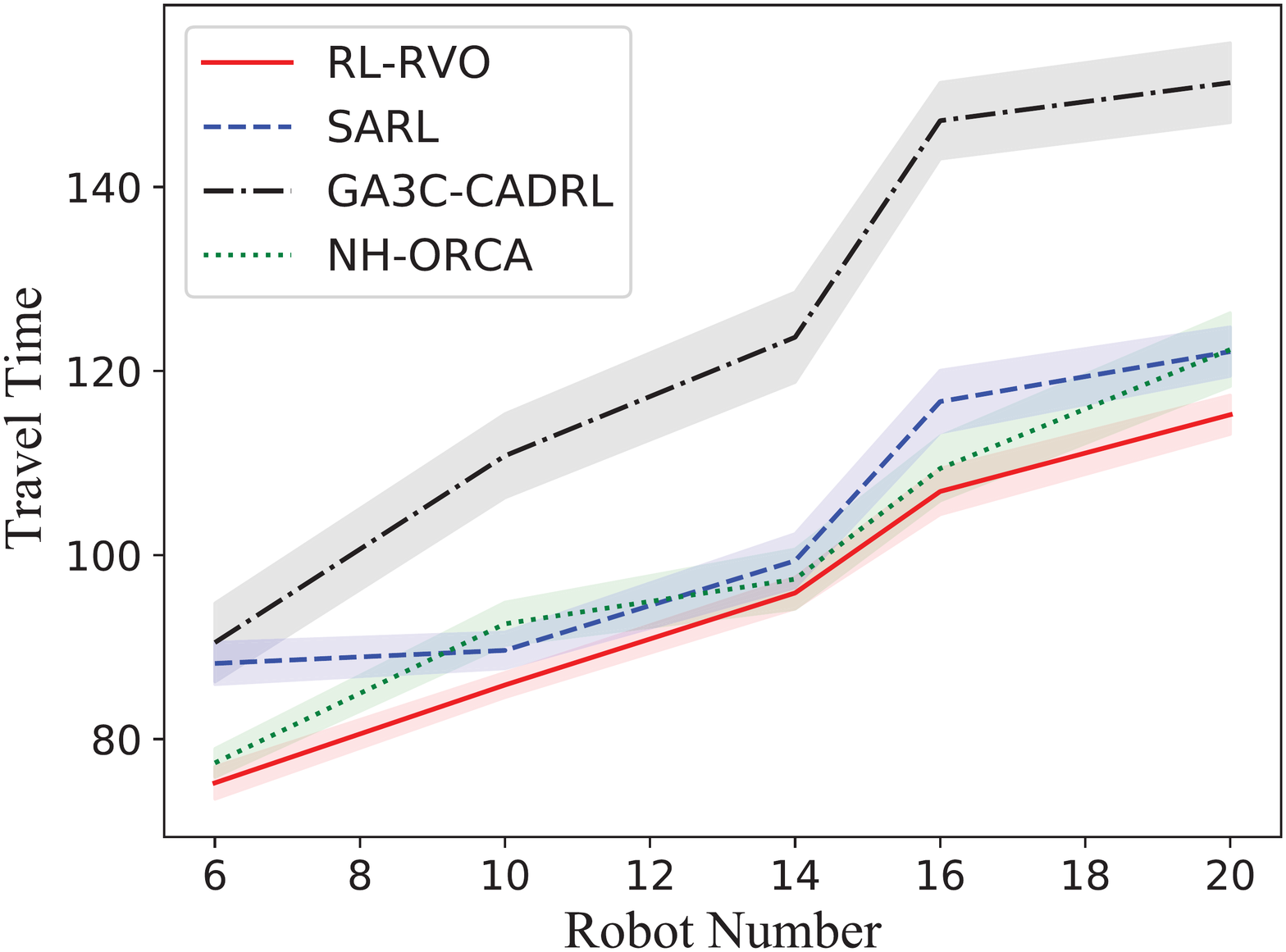}\hfill
        \includegraphics[width=0.33\linewidth]{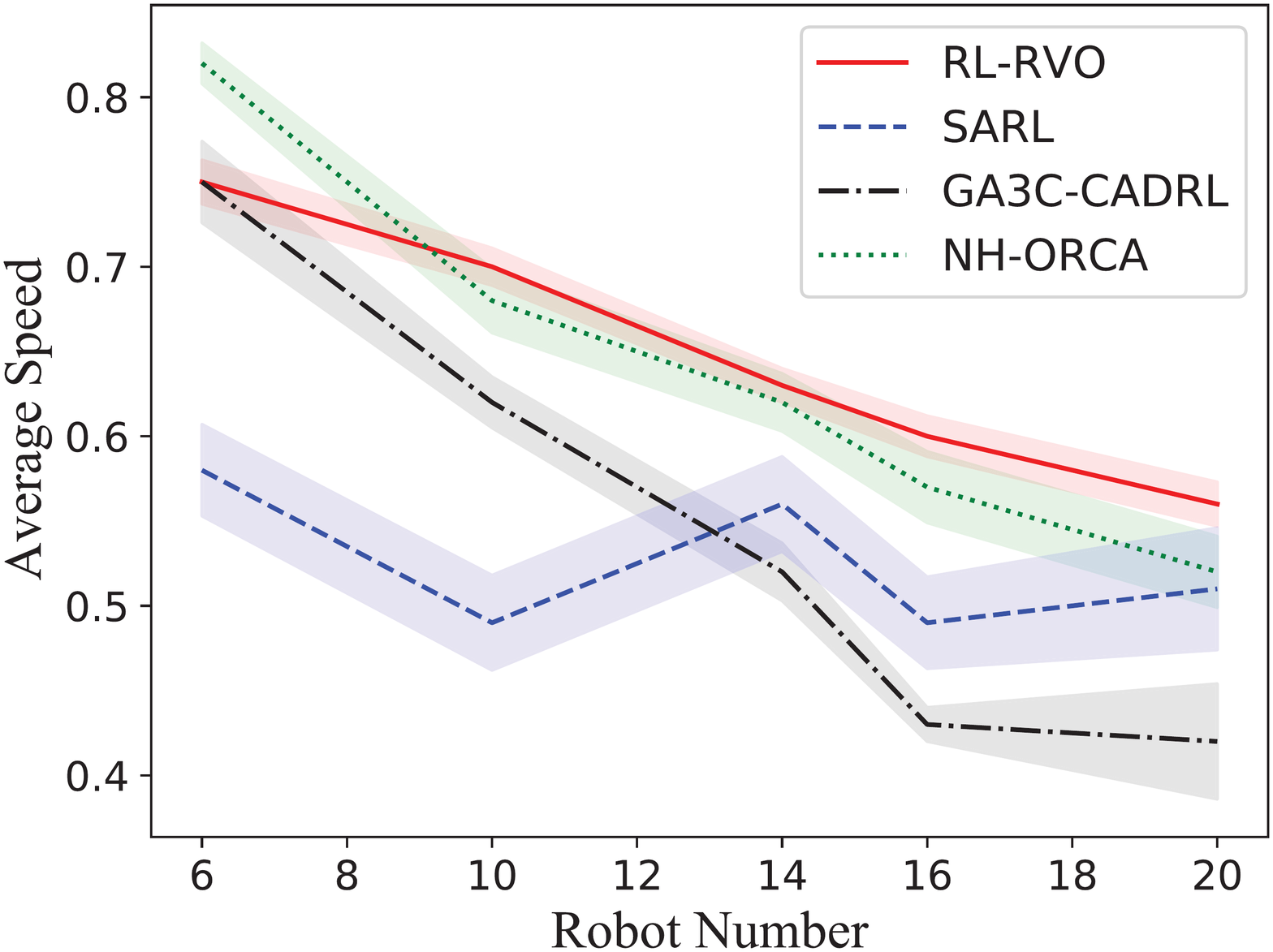}\hfill
        \subcaption{The results in the random scenario.}
    \end{subfigure}\hfill

    \caption{The testing results of four policies in the simulated scenarios in terms of three metrics (the standard deviation values are illustrated proportionally by the size of shades).}
    \label{results_fig}
    \vspace{-0.4cm}
\end{figure}

Generally, as the number of robots increases, the success rate, and average speed decrease, whereas the travel time increases. In the circle scenario, compared with the other three policies, our policy has the highest success rate, average speed, and lowest travel time, especially in the crowded scenario, as shown in Fig.~\ref{results_fig} (a). When there is a small number of robots, the policies perform comparably in the sparse scenario. As the number of robots increases, other policies have a heavy decrease in terms of the success rate, while our policy still works well. Similarly, in the randomly generated scenarios, our policy has better performance in various dense situations, as shown in Fig.~\ref{results_fig} (b). SARL and GA3C-CADRL predict the future state of the dynamic obstacles for decision making. However, the selection schemes of the control action are easy to encounter a freezing point problem, which influences the success rate in the random scenario. In comparison, our policy utilizes the BiGRU module to find more underlying relationships among surrounding robots and predict the future situation of the given limited information. Besides, the VO observation and reward guide the robots to achieve RCA behavior, which also improves the ability of collision avoidance and efficiency in dense situations. The results of the standard deviations also demonstrate the stability of our approach. In addition, the computational cost of our approach is lower than the other two learning based approaches a lot, which will influence the real-time performance of the policy. To summarize, compared with other policies, our policy is more robust and efficient in crowded situations under limited information.

\subsubsection{Ablation Study}

We also compare our policy with RL-NRVO, RL-LSTM and RL-Reward to demonstrate the functionality of components of our approach in the circle, random, and corridor scenarios, as shown in Fig.~\ref{scen}. All the policies are trained in the same DRL framework but with different components. RL-NRVO is the policy using the original information of robots as input instead of RVO vectors. This information includes relative positions/velocities and robot radiuses, as introduced in~\cite{han2020cooperative}, \textit{i.e.}, ${{\bf{o}}_{\text{sur}}} = [{p_x},{p_y},{v_x},{v_y},R]$. Similarly, RL-LSTM is the policy which replaces the BiGRUs with LSTM to tackle the input information of robots with varying numbers. And RL-Reward utilizes the distance-based reward function to train the policy under the same observations, as described in~\cite{long2018towards}. The results are listed in Table~\ref{ablation}. It can be seen that other policies can tackle the circle scenario well but have a dramatic performance decrease in the untrained (random and corridor) scenarios. The reason is that RL-NRVO does not learn an appropriate and regular behavior from the non-RVO input. Using LSTM to represent the robot information with variable numbers is not enough to tackle the crowded workspace. The distance base reward function is unmatched with the RVO based observations.

\begin{table*}[]
    \centering
    \caption{A performance comparison of four approaches in the circle and random scenarios with different numbers of robots.}
    \label{results}
    \resizebox{\textwidth}{!}{%
    \begin{tabular}{|ccccclcccclccccl|}
    \hline
    \multicolumn{1}{|c|}{\multirow{2}{*}{\begin{tabular}[c]{@{}c@{}}Robot\\ Number\end{tabular}}} & \multicolumn{5}{c|}{Success Rate}                                                                                                 & \multicolumn{5}{c|}{Travel Time (iteration step) / std}                                                                                                      & \multicolumn{5}{c|}{Average Speed (m/s) / std}                                                                                                    \\ \cline{2-16} 
    \multicolumn{1}{|c|}{}                                                                        & \multicolumn{1}{c|}{\textbf{RL-RVO}} & \multicolumn{1}{c|}{SARL} & \multicolumn{1}{c|}{GA3C-CADRL} & \multicolumn{2}{c|}{NH-ORCA} & \multicolumn{1}{c|}{\textbf{RL-RVO}}       & \multicolumn{1}{c|}{SARL}         & \multicolumn{1}{c|}{GA3C-CADRL}   & \multicolumn{2}{c|}{NH-ORCA}      & \multicolumn{1}{c|}{\textbf{RL-RVO}}    & \multicolumn{1}{c|}{SARL}      & \multicolumn{1}{c|}{GA3C-CADRL} & \multicolumn{2}{c|}{NH-ORCA}   \\ \hline
    \multicolumn{16}{|c|}{\textit{sensing range: 4m, circle scenario}}                                                                                                                                                                                                                                                                                                                                                                                                                                                                       \\ \hline
    \multicolumn{1}{|c|}{6}                                                                       & \multicolumn{1}{c|}{\textbf{1}}      & \multicolumn{1}{c|}{1}    & \multicolumn{1}{c|}{0.98}       & \multicolumn{2}{c|}{1}       & \multicolumn{1}{c|}{\textbf{78.29/4.48}}   & \multicolumn{1}{c|}{78.79/11.32}  & \multicolumn{1}{c|}{101.0/35.08}  & \multicolumn{2}{c|}{79.26/6.63}   & \multicolumn{1}{c|}{\textbf{1.09/0.06}} & \multicolumn{1}{c|}{1.08/0.12} & \multicolumn{1}{c|}{1.05/0.24}  & \multicolumn{2}{c|}{1.15/0.09} \\ \hline
    \multicolumn{1}{|c|}{10}                                                                      & \multicolumn{1}{c|}{\textbf{0.99}}   & \multicolumn{1}{c|}{0.89} & \multicolumn{1}{c|}{0.82}       & \multicolumn{2}{c|}{1}       & \multicolumn{1}{c|}{\textbf{90.23/4.16}}   & \multicolumn{1}{c|}{92.31/18.32}  & \multicolumn{1}{c|}{101.96/7.59}  & \multicolumn{2}{c|}{91.32/20.3}   & \multicolumn{1}{c|}{\textbf{0.98/0.04}} & \multicolumn{1}{c|}{0.92/0.18} & \multicolumn{1}{c|}{0.97/0.15}  & \multicolumn{2}{c|}{0.94/0.12} \\ \hline
    \multicolumn{1}{|c|}{14}                                                                      & \multicolumn{1}{c|}{\textbf{0.97}}   & \multicolumn{1}{c|}{0.82} & \multicolumn{1}{c|}{0.75}       & \multicolumn{2}{c|}{0.97}    & \multicolumn{1}{c|}{\textbf{103.13/5.04}}  & \multicolumn{1}{c|}{115.04/43.61} & \multicolumn{1}{c|}{125.7/27.98}  & \multicolumn{2}{c|}{116.25/16.59} & \multicolumn{1}{c|}{\textbf{0.88/0.04}} & \multicolumn{1}{c|}{0.91/0.24} & \multicolumn{1}{c|}{0.93/0.17}  & \multicolumn{2}{c|}{0.77/0.1}  \\ \hline
    \multicolumn{1}{|c|}{16}                                                                      & \multicolumn{1}{c|}{\textbf{0.93}}   & \multicolumn{1}{c|}{0.76} & \multicolumn{1}{c|}{0.61}       & \multicolumn{2}{c|}{0.89}    & \multicolumn{1}{c|}{\textbf{111.75/7.88}}  & \multicolumn{1}{c|}{128.85/38.52} & \multicolumn{1}{c|}{141.4/33.76}  & \multicolumn{2}{c|}{130.43/25.04} & \multicolumn{1}{c|}{\textbf{0.83/0.05}} & \multicolumn{1}{c|}{0.90/0.31} & \multicolumn{1}{c|}{0.86/0.1}   & \multicolumn{2}{c|}{0.72/0.12} \\ \hline
    \multicolumn{1}{|c|}{20}                                                                      & \multicolumn{1}{c|}{\textbf{0.90}}   & \multicolumn{1}{c|}{0.71} & \multicolumn{1}{c|}{0.34}       & \multicolumn{2}{c|}{0.80}    & \multicolumn{1}{c|}{\textbf{128.62/9.52}}  & \multicolumn{1}{c|}{138.73/33.05} & \multicolumn{1}{c|}{152/30.23}    & \multicolumn{2}{c|}{143.51/27.89} & \multicolumn{1}{c|}{\textbf{0.76/0.07}} & \multicolumn{1}{c|}{0.80/0.19} & \multicolumn{1}{c|}{0.75/0.34}  & \multicolumn{2}{c|}{0.70/0.13} \\ \hline
    \multicolumn{16}{|c|}{\textit{sensing range: 4m, random scenario}}                                                                                                                                                                                                                                                                                                                                                                                                                                                                       \\ \hline
    \multicolumn{1}{|c|}{6}                                                                       & \multicolumn{1}{c|}{\textbf{1}}      & \multicolumn{1}{c|}{0.93} & \multicolumn{1}{c|}{0.81}       & \multicolumn{2}{c|}{0.94}    & \multicolumn{1}{c|}{\textbf{75.24/17.59}}  & \multicolumn{1}{c|}{88.22/22.83}  & \multicolumn{1}{c|}{90.49/52.73}  & \multicolumn{2}{c|}{77.41/15.49}  & \multicolumn{1}{c|}{\textbf{0.75/0.13}} & \multicolumn{1}{c|}{0.58/0.27} & \multicolumn{1}{c|}{0.75/0.24}  & \multicolumn{2}{c|}{0.82/0.12} \\ \hline
    \multicolumn{1}{|c|}{10}                                                                      & \multicolumn{1}{c|}{\textbf{0.98}}   & \multicolumn{1}{c|}{0.89} & \multicolumn{1}{c|}{0.72}       & \multicolumn{2}{c|}{0.92}    & \multicolumn{1}{c|}{\textbf{85.88/13.49}}  & \multicolumn{1}{c|}{89.64/19.78}  & \multicolumn{1}{c|}{110.79/45.53} & \multicolumn{2}{c|}{92.53/23.38}  & \multicolumn{1}{c|}{\textbf{0.70/0.11}} & \multicolumn{1}{c|}{0.49/0.28} & \multicolumn{1}{c|}{0.62/0.15}  & \multicolumn{2}{c|}{0.68/0.19} \\ \hline
    \multicolumn{1}{|c|}{14}                                                                      & \multicolumn{1}{c|}{\textbf{0.97}}   & \multicolumn{1}{c|}{0.71} & \multicolumn{1}{c|}{0.61}       & \multicolumn{2}{c|}{0.84}    & \multicolumn{1}{c|}{\textbf{95.88/17.12}}  & \multicolumn{1}{c|}{99.38/29.29}  & \multicolumn{1}{c|}{123.67/48.93} & \multicolumn{2}{c|}{97.38/32.55}  & \multicolumn{1}{c|}{\textbf{0.63/0.1}}  & \multicolumn{1}{c|}{0.56/0.28} & \multicolumn{1}{c|}{0.52/0.17}  & \multicolumn{2}{c|}{0.62/0.17} \\ \hline
    \multicolumn{1}{|c|}{16}                                                                      & \multicolumn{1}{c|}{\textbf{0.96}}   & \multicolumn{1}{c|}{0.69} & \multicolumn{1}{c|}{0.42}       & \multicolumn{2}{c|}{0.70}    & \multicolumn{1}{c|}{\textbf{106.91/25.1}}  & \multicolumn{1}{c|}{116.68/33.85} & \multicolumn{1}{c|}{147.2/41.49}  & \multicolumn{2}{c|}{109.41/35.18} & \multicolumn{1}{c|}{\textbf{0.60/0.12}} & \multicolumn{1}{c|}{0.49.0.27} & \multicolumn{1}{c|}{0.43/0.1}   & \multicolumn{2}{c|}{0.57/0.21} \\ \hline
    \multicolumn{1}{|c|}{20}                                                                      & \multicolumn{1}{c|}{\textbf{0.92}}   & \multicolumn{1}{c|}{0.60} & \multicolumn{1}{c|}{0.36}       & \multicolumn{2}{c|}{0.65}    & \multicolumn{1}{c|}{\textbf{115.25/21.22}} & \multicolumn{1}{c|}{122.11/26.72} & \multicolumn{1}{c|}{151.33/43.21} & \multicolumn{2}{c|}{122.33/39.96} & \multicolumn{1}{c|}{\textbf{0.56/0.13}} & \multicolumn{1}{c|}{0.51/0.36} & \multicolumn{1}{c|}{0.42/0.34}  & \multicolumn{2}{c|}{0.52/0.21} \\ \hline
    \end{tabular}%
    }
    \vspace{-0.5cm}
    \end{table*}


\begin{table}[t]
    \centering
    \caption{The results of ablation experiments}
    \label{ablation}
    \resizebox{0.47\textwidth}{!}{%
    \begin{tabular}{|c|ccc|}
    \hline
    \textit{Setup}  & \multicolumn{3}{c|}{\textit{\begin{tabular}[c]{@{}c@{}}robot number 8, sensing range: 4m, \\ circle/random/corridor scenario\end{tabular}}} \\ \hline
    Policy          & \multicolumn{1}{c|}{Success Rate}               & \multicolumn{1}{c|}{Travel Time}                        & Average Speed             \\ \hline
    \textbf{RL-RVO} & \multicolumn{1}{c|}{\textbf{1/1/0.89}}          & \multicolumn{1}{c|}{\textbf{83.75/79.71/84.23}}         & \textbf{1.04/0.73/0.89}         \\ \hline
    RL-NRVO         & \multicolumn{1}{c|}{0.97/0.53/0.29}             & \multicolumn{1}{c|}{94.45/85.22/89.31}                  & 0.94/0.70/0.84                  \\ \hline
    RL-LSTM         & \multicolumn{1}{c|}{0.96/0.85/0.42}             & \multicolumn{1}{c|}{90.05/91.65/85.43}                  & 0.96/0.69/0.70                  \\ \hline
    RL-Reward       & \multicolumn{1}{c|}{0.93/0.71/0.58}             & \multicolumn{1}{c|}{98.54/88.87/92.61}                  & 0.91/0.68/0.67                  \\ \hline
    \end{tabular}%
    }
    \end{table}

\begin{table}[t]
    \centering
    \caption{Results of Real-World Experiment}
    \label{result_real}
    \resizebox{0.5\textwidth}{!}{%
    \begin{tabular}{|c|c|c|c|c|c|c|}
    \hline
    \textit{Test Setup} & \multicolumn{6}{c|}{\textit{realistics scenario}}                                                         \\ \hline
    Robot Number        & \multicolumn{2}{c|}{Success Rate} & \multicolumn{2}{c|}{Travel Time (s) / std} & \multicolumn{2}{c|}{Average Speed (m/s) / std} \\ \hline
                        & \textbf{RL-RVO}     & NH-ORCA     & \textbf{RL-RVO}         & NH-ORCA      & \textbf{RL-RVO}      & NH-ORCA     \\ \hline
    4                   & \textbf{1}          & 1           & \textbf{13.96/1.09}     & 14.98/1.16  & \textbf{0.81/0.11}    & 0.79/0.21        \\ \hline
    6                   & \textbf{0.91}       & 0.85        & \textbf{16.01/0.92}     & 18.45/1.11    & \textbf{0.74/0.12}  & 0.72/0.17        \\ \hline
    8                   & \textbf{0.89}       & 0.68        & \textbf{22.18/1.25}     & 28.94/1.57    & \textbf{0.69/0.15}  & 0.65/0.23        \\ \hline
    \end{tabular}%
    }
    \end{table}

\subsubsection{Real-world experiments}
\begin{figure}[tp]
    \centering
    \includegraphics[width=0.30\textwidth, clip]{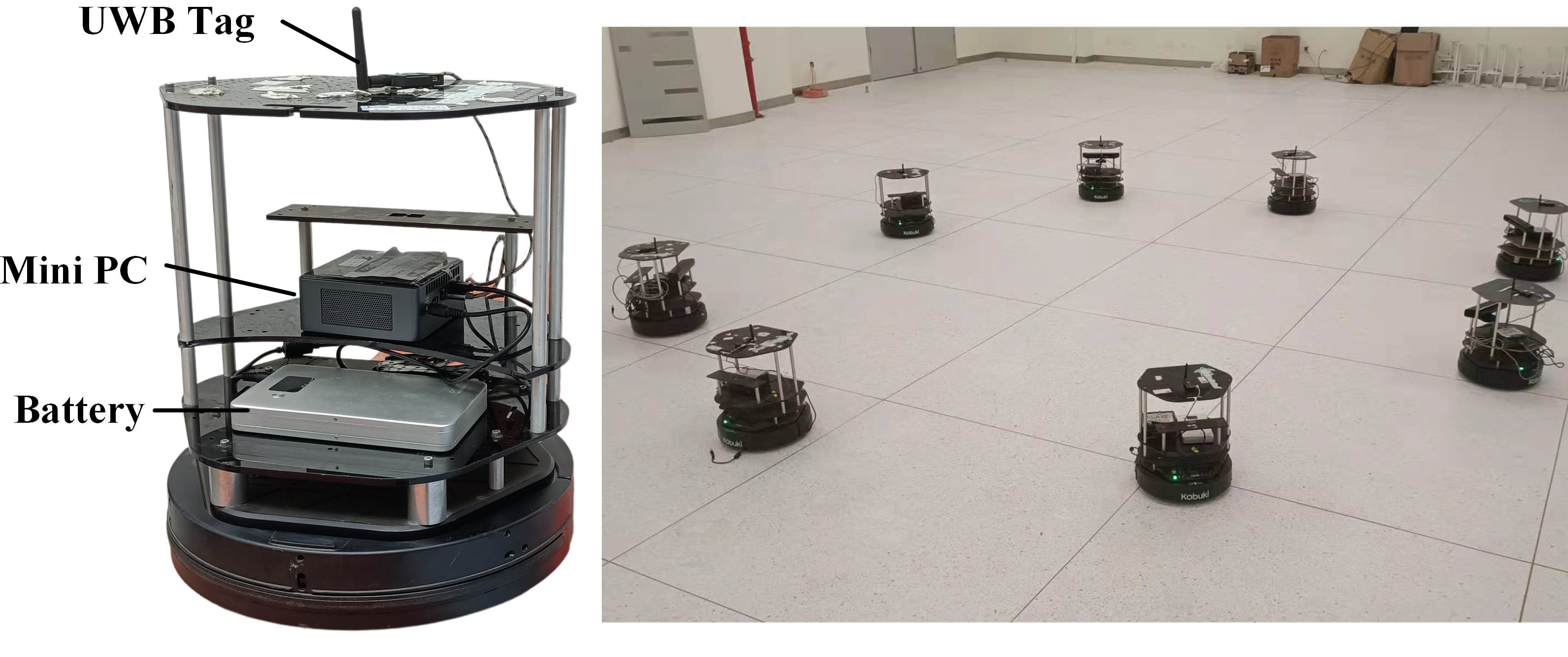}
    \caption{An illustration of real-world experiments: (left) a single Turtlebot, (right) eight Turtlebots positioned along a circle uniformly with random orientations.}
    \label{exp}
\end{figure}

The RL-RVO policy is implemented and tested in Turtlebot robots to demonstrate the performance of our policy in the real world. Those experiments use up to $8$ differential drive Turtlebots, as shown in Fig.~\ref{exp}. All the Turtlebots are arranged along a circle with random orientations. Each Turtlebot is equipped with a mini PC to compute velocity and a tag to localize itself by the UWB localization system. The robots receive the exteroceptive measurements through the ROS architecture. Specifically, due to the mechanical limit, the maximum velocity of each robot is set to be $1m/s$.

Because of the high computational cost of GA3C-CARDL and SARL, we only compare our policy with NH-ORCA using 4, 6, and 8 Turtlebots in the real world experiment with circle scenario. Statistics across 50 cases are described in Table~\ref{result_real}. Different from simulated experiments, there are ubiquitous uncertainties in real-world experiments. NH-ORCA does not take noise and uncertainties into account, thus its performance is less robust in real-world experiments. In contrast, our policy has a better success rate, and it also takes less travel time to accomplish the same navigation tasks than NH-ORCA.

\section{Conclusion}\label{section 7}
In this paper, we have presented a DRL based multi-robot navigation approach, which utilizes RVO techniques to tackle collision avoidance problems. The proposed RVO based dynamic and static environment state representation can better describe the reciprocal interactions. The developed BiGRUs based neural network can extract environment features despite the varying number of moving robots and map the features to the control actions directly with low cost. The RVO area and expected collision time based reward function can achieve reciprocal collision behaviors and a trade-off between the collision risk and travel time. Both simulated and real-world experiments have been performed respectively to evaluate the policy's navigation performance. Four policies, including RL-RVO, SARL, GA3C-CADRL, and NH-ORCA have been compared in terms of success rate, travel time, and average speed in both circle and random scenarios. The experiment results have shown that the proposed approach has a superior collision avoidance capability and time efficiency over other methods in the crowded environment. The ablation study demonstrates the functionality of the individual components in our approach. Furthermore, it has been shown that our policy has been generalized well to tackle situations with more numbers of robots than those used for training. Our future work includes extending such an approach to addressing multi-robot navigation problems with more challenging uncertainties in the real world.

\bibliographystyle{IEEEtran}
\bibliography{reference/Thesis}

\end{document}